# SPOT!: Map-Guided LLM Agent for Unsupervised Multi-CCTV Dynamic Object Tracking

Yujin Roh[ǂ]
*Artificial Intelligence Engineering*
*College of Software and Convergence,*
*Kwangwoon University*
Seoul, South Korea
rohyujin55@gmail.com
0009-0001-5706-6129

Inho Jake Park[ǂ]
*Mechanical Engineering*
*Gwangju Institute of Science And Technology (GIST)*
GwangJu, South Korea
jakepark5848@gm.gist.ac.kr
0009-0004-1107-1274

Chigon Hwang[*]
*Computer Engineering*
*Institute of Information Technology*
*Kwangwoon University*
Seoul, South Korea
duck1052@kw.ac.kr
0000-0002-3403-540X

***Abstract*****—** CCTV-based vehicle tracking systems face structural limitations in continuously connecting the trajectories of the same vehicle across multiple camera environments. In particular, blind spots occur due to the intervals between CCTVs and limited Fields of View (FOV), which leads to object ID switching and trajectory loss, thereby reducing the reliability of real-time path prediction. This paper proposes SPOT (Spatial Prediction Over Trajectories), a map-guided LLM agent capable of tracking vehicles even in blind spots of multi-CCTV environments without prior training. The proposed method represents road structures (Waypoints) and CCTV placement information as documents based on 2D spatial coordinates and organizes them through chunking techniques to enable real-time querying and inference. Furthermore, it transforms the vehicle's position into the actual world coordinate system using the relative position and FOV information of objects observed in CCTV images. By combining map spatial information with the vehicle's moving direction, speed, and driving patterns, a beam search is performed at the intersection level to derive candidate CCTV locations where the vehicle is most likely to enter after the blind spot. Experimental results based on the CARLA simulator in a virtual city environment confirmed that the proposed method accurately predicts the next appearing CCTV even in blind spot sections, maintaining continuous vehicle trajectories more effectively than existing techniques.

## I. INTRODUCTION

Geospatial image registration is a fundamental technology for synthesizing a comprehensive global map from heterogeneous visual data acquired by various sensors, including CCTVs, drones, and autonomous robots. Conventional approaches typically perform registration by establishing correspondences between feature points extracted from images to link different viewpoints. However, in practical urban deployments, wide camera intervals and restricted Fields of View (FOV) often prevent fixed landmarks from being captured simultaneously across multiple frames. These constraints hinder the application of traditional registration methods, leading to severe performance degradation in both object tracking and spatial localization—especially within "blind spots" where visual overlap between cameras is absent. To address these challenges, this study introduces **SPOT! (Spatial Prediction Over Trajectories)**, a multi-CCTV blind-spot tracking framework driven by a map-guided Large Language Model (LLM) agent. As illustrated in Fig. 1, the framework performs indirect geospatial registration by integrating observed object trajectories with structural map data to infer possible movement paths across visually disconnected segments. The primary innovation of SPOT lies in enabling zero-shot spatial reasoning for LLMs by structuring map information as textual documents. Road structures (waypoints) and CCTV configurations are represented as 2D coordinate-based documents and chunked by intersection to support real-time querying. These documents serve as the knowledge base for a Retrieval-Augmented Generation (RAG) system. Vehicle positions are projected onto the global map by transforming pixel-level coordinates into the world coordinate system using intrinsic and extrinsic camera parameters. For tracking through blind spots, the framework employs a beam search algorithm that evaluates traversable paths based on map connectivity, vehicle velocity, heading, and historical driving patterns. This process derives the Top-K most probable next CCTV locations, with the final selection refined by a driving pattern scoring mechanism. The proposed system was validated using a virtual urban environment built on the CARLA simulator. Realistic parameters—including camera height, pitch, and FOV—were utilized to ensure parity with real-world infrastructure. Furthermore, we implemented a RAG-based diagnostic system to evaluate the agent's spatial understanding and provide explainable rationales for inferred trajectories, blind-spot reasoning, and route recommendations. The remainder of this paper is organized as follows: **Section II** reviews related works. **Section III** describes the architecture of the proposed framework. Section IV details the experimental setup and performance analysis. **Section V** summarizes the findings and discusses future research directions.

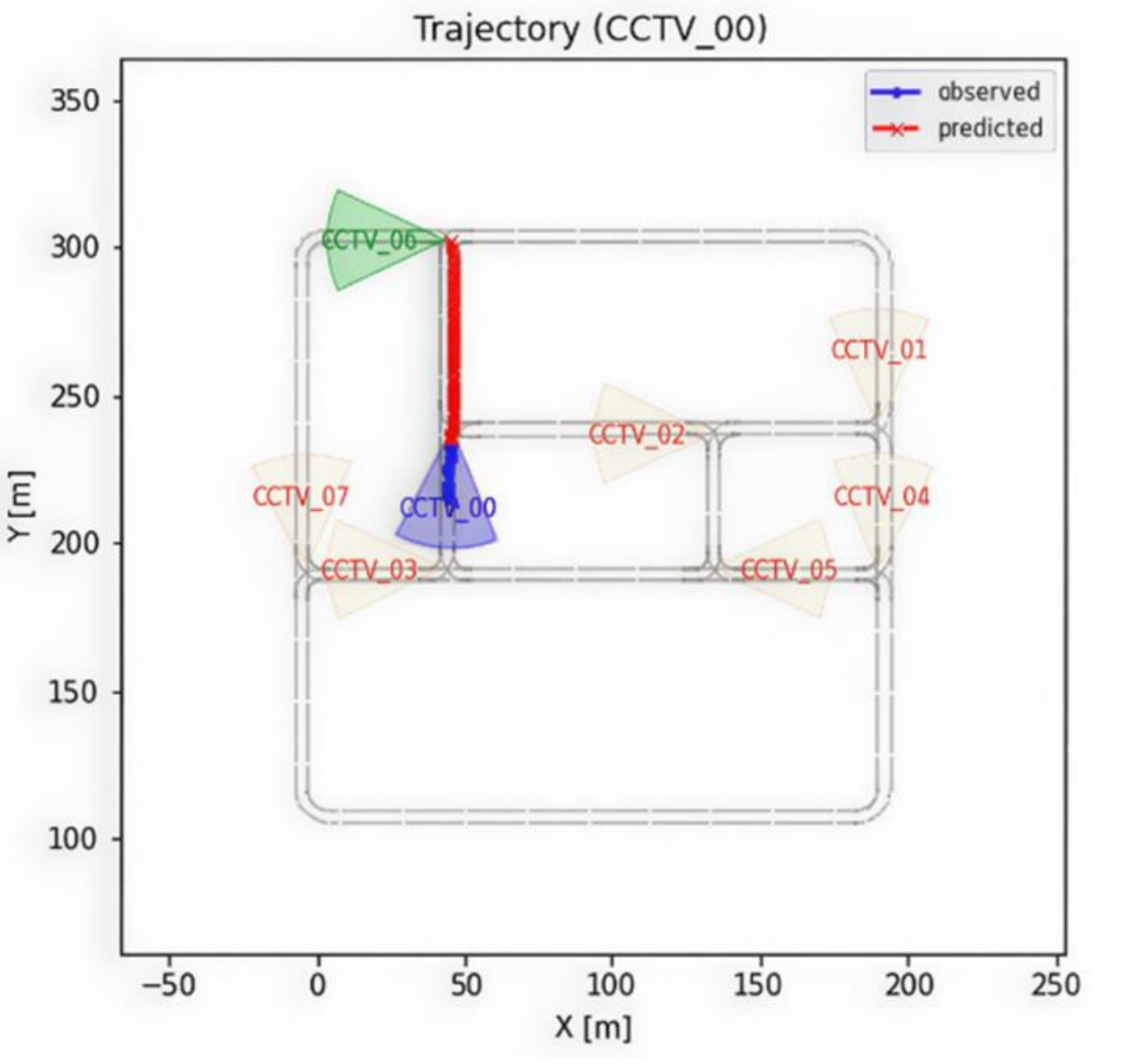

Fig 1. Example of subsequent CCTV estimation via vehicle trajectory prediction within blind spots outside the Field of View (FOV)

ǂ: These authors contributed equally to this work. The authors are listed in alphabetical order by their last names.
Under consideration at Pattern Recognition Letters

## II. Related Works

### 2.1. Object Tracking

Object tracking is a technique for estimating the position changes of an object while maintaining its identity across consecutive video frames, typically performed based on object detection results. It plays a critical role in various applications such as surveillance systems, autonomous driving, and robot vision by enabling continuous identification of dynamic objects or analysis of their movement paths. Conventional research has primarily focused on enhancing tracking accuracy in single-camera environments, evolving through state estimation techniques like Kalman filters and particle filters, or data association methods utilizing appearance features. With the recent introduction of deep learning-based tracking, approaches that learn object appearance representations to improve ID maintenance performance have become widely adopted.

Tracking in multi-CCTV environments is significantly more challenging than in single-camera setups. Issues such as viewpoint differences, illumination changes, FOV inconsistencies, and the presence of blind spots frequently lead to ID switches and trajectory fragmentation. While multi-camera tracking methods using appearance matching or spatio-temporal constraints have been proposed to mitigate these issues, they face clear limitations in environments where there is no visual overlap between cameras. Furthermore, most existing vehicle tracking research is confined to trajectory analysis within the image coordinate system. There has been limited exploration of world coordinate transformation for interpreting paths in real-world road environments or reasoning that considers road structures and high-dimensional behavioral information, such as driver habits and movement patterns. This study proposes a tracking framework that stably tracks vehicles in CCTV videos, transforms these positions into world coordinate trajectories on a map, and integrates this information with driver pattern analysis to infer paths even in visually disconnected blind spots.

### 2.2. Large Language Model (LLM)

Large Language Models (LLMs) are autoregressive language models pre-trained on massive text datasets. They have secured general-purpose question-answering capabilities through model scaling and instruction-following alignment. In particular, few-shot learning and Reinforcement Learning from Human Feedback (RLHF) have significantly improved generalization performance for new tasks. Additionally, Chain-of-Thought (CoT) prompting and reasoning-action coupling techniques enable step-by-step reasoning and tool utilization for complex problems.

In the context of map and geographic information, LLMs transform structured spatial information into linguistic expressions to perform question-answering and reasoning. Related research follows three main trends: first, spatial question-answering (QA) that connects geographic knowledge with linguistic queries; Spatial-RAG studies have reported performance gains in urban and tourism queries by augmenting text contexts with coordinates, place names, routes, and adjacency relationships. Second, approaches combining geographic knowledge graphs with LLMs, such as GeoRAG and GeoGraphRAG, retrieve subgraphs for LLMs to interpret and generate explanatory responses. In this study, the LLM serves two roles: first, as a Map QA Engine, it organizes road, intersection, and CCTV layout information into text contexts to respond to spatial queries. Here, RAG retrieves spatial candidates, and the LLM handles linguistic integration, such as route descriptions and rationale summarization. Second, as a Local Re-ranker, it evaluates context differences and adjusts scores between top candidates during intersection branching and next-CCTV selection steps. In this capacity, the LLM does not generate new candidates but performs reasoning within the given structural constraints.

### 2.3. Retrieval-Augmented Generation (RAG)

Retrieval-Augmented Generation (RAG) is a framework that integrates external knowledge retrieval into the generation process, showing effective performance in knowledge-intensive tasks. Lewis et al. systematized large-scale knowledge retrieval and text generation by proposing a RAG structure combining non-parametric memory (vector indices) with generative models. Representative vector search infrastructures include FAISS and ScaNN, which support large-scale search through approximate nearest neighbor search and high-speed inner product calculations.

In cases where embedding training is difficult or robust search against domain notation variations is required, non-learning-based techniques—such as approximating Jaccard similarity based on token k-shingles using MinHash—are utilized for large-scale similar document retrieval. MinHash offers high operational efficiency as it allows error control through signature length. In map and geographic information fields, spatial keyword search combining spatial and text indices is standard. IR-tree structures integrate keyword summaries into R-tree nodes to prune spatial and text conditions simultaneously, while S2I accelerates top-k spatial queries using keyword-based aggregate R-trees. Recently, Spatial RAG, which jointly considers spatial constraints and semantic relevance, has been proposed with reported effectiveness in urban and tourism QA. Furthermore, extension techniques linking subgraph retrieval with LLM reasoning by incorporating geographic knowledge graphs are emerging. Existing research has evolved from (i) embedding-based general RAG to (ii) spatial keyword search combining spatial pruning and semantic re-ranking, and (iii) Spatial RAG jointly optimizing spatial constraints and semantics. Based on this progression, this paper adopts a structural constraint-based integrated configuration utilizing spatial indexing (R-tree) and non-learning semantic approximation (MinHash) within public and commercial map API environments.

## III. Method

This section details the methodology of the proposed **SPOT! (Spatial Prediction Over Trajectories)** framework, designed to perform vehicle tracking across multiple CCTV cameras including blind-spot sections. As illustrated in Fig. 2, SPOT transforms vehicle trajectories obtained from CCTV video into a world coordinate system. By leveraging map-based spatial information, it infers movement paths even in visually disconnected segments to predict the most probable subsequent CCTV. The core of the proposed method focuses on resolving trajectory fragmentation between multiple CCTVs by combining structured map documents with LLM-based spatial reasoning, without relying on external training data.

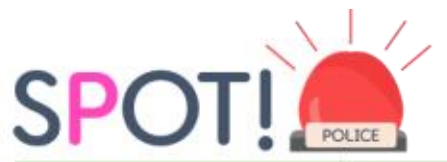

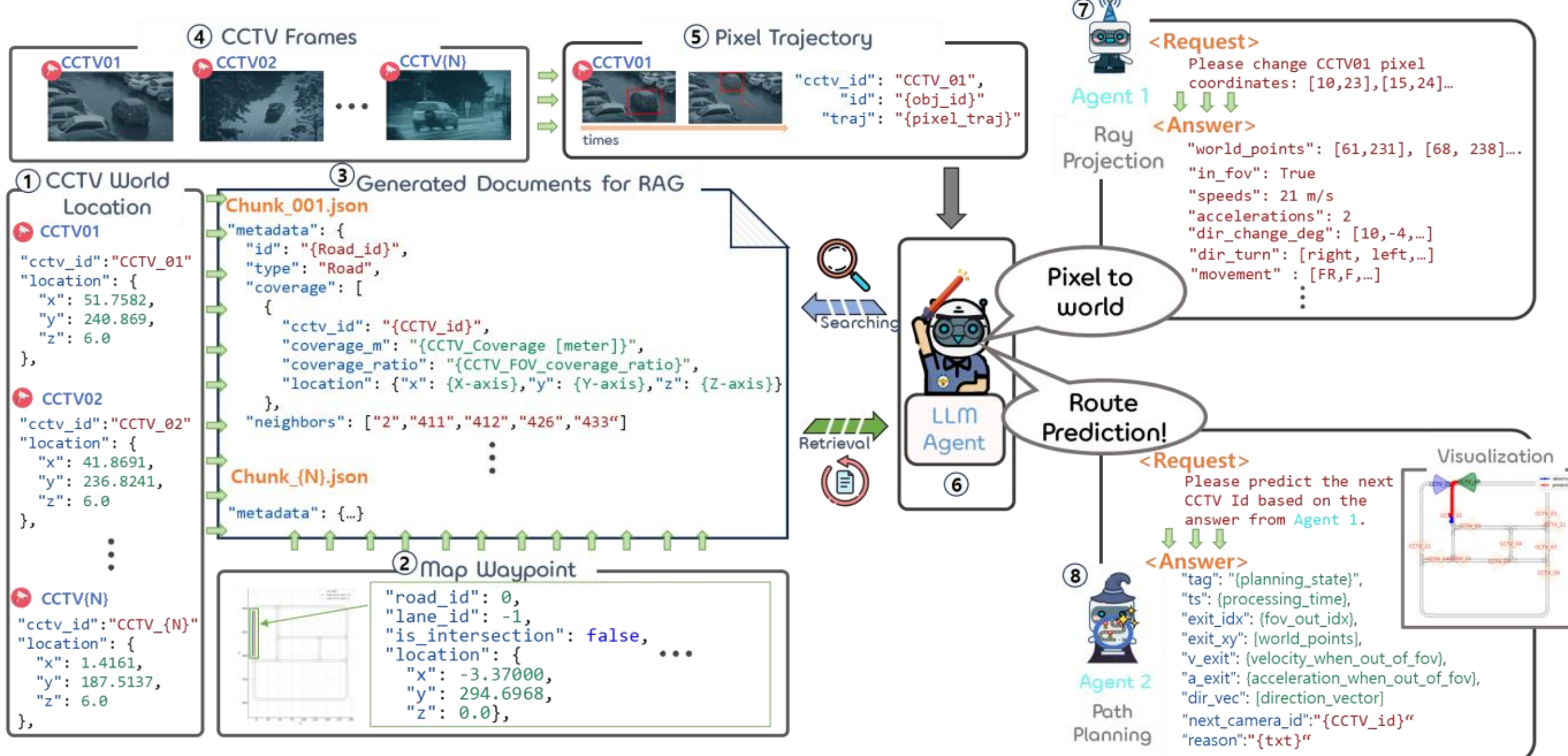

Fig 2. (1) the location and field-of-view parameters of each CCTV camera; (2) a map representation incorporating road topology and intersection structure; (3) generation of structured map documents to enable spatial reasoning by the language model; (4) video frames acquired from CCTV streams; (5) the target vehicle trajectory and the current CCTV context obtained via object detection and tracking; (6) the SPOT agent module performing map-based spatial reasoning; (7) pixel-to-world coordinate transformation within the current CCTV field of view performed by Agent 1; and (8) route inference and selection of feasible next CCTV cameras performed by Agent 2.

### *3.1 .Configuration of LLM Agent for Map-based Documentation and Question-Answering*

We designed a map-based LLM agent framework to solve visibility disconnection issues and ensure continuous vehicle tracking. The framework operates through a **Semantic Orchestration** process led by an LLM orchestrator. When the orchestrator receives vehicle trajectories detected in pixel coordinates from a CCTV, it first calls **Agent 1 (Perception Agent)** to perform the pixel-to-world coordinate transformation and extract numerical kinematic states (position, velocity, heading). This processed physical information is then passed to **Agent 2 (Planning Agent)** for high-level spatial reasoning and path prediction. To support this orchestration, we propose a map documentation technique that converts actual map data—based on CCTV metadata and road structures provided in (1) and (2) of Fig. 2—into a structured document set (3). The documentation process involves:

**Road Representation**: Waypoints are extracted from road centerline data at approximately 1m intervals, with each segment defined as a document unit containing ID, location, and yaw.

**Network Topology**: The entire road network is expressed through segment connectivity, with intersections classified as CCTV candidate regions via clustering.

**CCTV Modeling**: Metadata (position, direction, FOV, distance) are used to construct FOV polygons representing surveillance ranges.

**Road-CCTV Gates**: Pre-analyzed intersections between FOV polygons and waypoints define **"Road-CCTV Gates,"** which allow Agent 1 to trigger the handover by detecting when a vehicle exits a camera's view.

The entire map is composed of a collection of individual documents, following the format:

| *Road 131: Waypoints = 18, Coverage = CCTV_13 (54.3%), CCTV_00 (12.8%), Neighbors = [8, 11, 92, 93]* |
|---|

The map's knowledge base is managed via an R-tree spatial index and MinHash signatures. During the **Semantic Orchestration** process, once the orchestrator receives the kinematic states from Agent 1, it initiates **VLA Tokenization** to transform these continuous physical trajectories into discrete semantic tokens. Subsequently, **Agent 2** executes **Hybrid Retrieval** (Spatial/Semantic filtering) to extract the most relevant map context.

Finally, the orchestrator synthesizes a **Physics-informed prompt** by integrating the retrieved map documents with the serialized trajectory predictions. This allows Agent 2 to perform spatial reasoning, providing natural language-based QA and explainable rationales for predicting movement paths in blind spots and selecting subsequent CCTV candidates.

**Algorithm 1** Semantic Orchestration

**Require:** Query $Q$, State $\mathcal{S}_t$, Path $P^*$, Indices $\mathcal{I}$
**Ensure:** Response $R$
1: **procedure** GENERATERESPONSE($Q, \mathcal{S}_t, P^*$)
2: *// 1. VLA Tokenization*
3: $T_{scene} \leftarrow$ "[SCENE] pos=" $+ \mathcal{S}_t$.pos $+ \ldots$
4: $\vec{v}_{scene} \leftarrow$ Embedding($T_{scene}$)
5: *// 2. Hybrid Retrieval*
6: $S_{spatial} \leftarrow \mathcal{I}_{spatial}$.Query($\mathcal{S}_t$.pos, $Rad$)
7: $S_{semantic} \leftarrow \{d \mid$ Jaccard(MH($Q$), $d$.MH) $> \tau\}$ ▷ MinHash Filter
8: $\mathcal{C}\dashv\setminus\lceil\int \leftarrow S_{spatial} \cap S_{semantic}$
9: *// 3. VLA Reranking*
10: $\mathcal{R}\dashv\setminus\|\rceil\lceil \leftarrow []$
11: **for** each Doc $d$ in $\mathcal{C}\dashv\setminus\lceil\int$ **do**
12: $T_{doc} \leftarrow$ BUILDTOKEN($d$)
13: $\vec{v}_{doc} \leftarrow$ Embed($T_{doc}$)
14: $P_{align} \leftarrow \sigma(\vec{v}_{scene} \cdot \vec{v}_{doc})$
15: $S_{fin} \leftarrow (1-\beta)d$.Base $+ \beta P_{align}$
16: $\mathcal{R}\dashv\setminus\|\rceil\lceil$.append(($d, S_{fin}$))
17: **end for**
18: $\mathcal{C}\sqcup\S \leftarrow$ TOPK($\mathcal{R}\dashv\setminus\|\rceil\lceil$, 5)
19: *// 4. Synthesis*
20: $Prt \leftarrow Q + \mathcal{C}\sqcup\S$ + Serialize($P^*$)
21: **return** LLM($Prt$)
22: **end procedure**

Fig 3. Pseudo Algorithm for Semantic Orchestration (LLM Agent)

Fig 3 describes the process of **Semantic Orchestration**. This agent serves as a bridge between the numerical physics engine and the Large Language Model (LLM), taking the user query ($q$), the current physical state ($s$), and the predicted optimal trajectory ($\tau$) as inputs.

*Step 1: VLA Tokenization (Lines 3–4)*

The algorithm first transforms continuous physical states into discrete semantic representations.

- **Line 3**: A structured state token ($T_s$ is generated in a text format (e.g., "[SCENE] pos=...") by combining the vehicle's position, movement history, and camera identifier.
- **Line 4**: This token is converted into a high-dimensional vector ($e$) using a pre-trained embedding model. This enables the comparison of numerical and textual data within the same vector space.

*Step 2: Hybrid Retrieval (Lines 6–8)*

To efficiently retrieve relevant information from a large-scale map database, a two-stage filtering strategy is employed.

- **Line 6**: A **Spatial Filter** utilizing an R-Tree index retrieves a set of documents ($D_s$ within a specific radius of the vehicle's position.
- **Line 7**: A **Semantic Filter** utilizing MinHash estimates the Jaccard similarity between the query ($q$) and the documents, selecting documents ($D_m$ that exceed a predefined similarity threshold ($\tau_{sim}$).
- **Line 8**: The initial candidate set ($D_{cand}$) is formed by taking the intersection of the two sets ($D_s \cap D_m$).

*Step 3: VLA Reranking (Lines 11–17)*

The selected candidate documents are reranked by evaluating their alignment with the current physical context (e.g., performing a right turn or a sudden stop).

- **Line 12**: A document token ($T_d$) is generated for each candidate document ($d$).
- **Line 14**: The **alignment probability** ($P_{align}$) is calculated by applying a sigmoid function to the dot product (or cosine similarity) between the state vector ($e$) and the document vector ($v$). This represents how well the static map data aligns with the dynamic vehicle state.
- **Line 15**: The final score ($S_{final}$) is derived by combining the initial retrieval score with the alignment probability using a dynamic weight ($\alpha$).

*Step 4: Synthesis (Lines 20–21)*

- **Line 20**: A **physics-informed prompt** ($X_{prompt}$) is constructed by integrating the top-ranked documents ($D_{top}$) with the serialized trajectory prediction results ($\tau$).
- **Line 21**: Finally, the LLM is invoked to generate a natural language response for the user.

### *3.2 World Coordinate Transformation of Object Positions Based on CCTV Field of View*

For the agent to analyze the trajectory of a tracked object, it is necessary to transform the object's position observed in pixel units within each CCTV image into the world coordinate system—the actual coordinate system of the map. This section details the procedure for projecting the 2D image coordinates of a vehicle, obtained from CCTV video, into 3D world coordinates, corresponding to step (7) in Fig. 2.

The transformation process is performed based on a camera projection model, utilizing both intrinsic and extrinsic parameters. The intrinsic parameter matrix K is composed of factors such as camera resolution, focal length, sensor size, and field of view (FOV); it transforms image coordinates into the camera coordinate system. The extrinsic parameters $[R|t]$ are configured based on the location and orientation of each CCTV defined in the simulator environment, handling the transformation between the world coordinate system and the camera coordinate system. Here, the rotation matrix $R$ represents the camera's orientation, and the translation vector $t$ indicates the position of the camera center. These parameters are pre-extracted for each CCTV unit within the simulator, and the parameter set $(K, R, t)$ is stored in a local cache to facilitate real-time access and enhance transformation performance. Object position transformation is executed based on ray-plane intersection analysis. The pixel coordinates $(u, v)$ of an object in the image are converted into a direction vector $\vec{d}_{cam}$ in the camera coordinate system using the inverse of the intrinsic parameter matrix $K^{-1}$, as follows:

$$\vec{d}_{cam} = K^{-1}\begin{bmatrix} u \\ v \\ 1 \end{bmatrix}$$

This direction vector is subsequently transformed into a world coordinate direction vector $d_{world}$ by applying the extrinsic parameters. By estimating the intersection point where this ray meets the ground (road plane, e.g., $z = 0$), image coordinates are mapped to physical locations on the map. This process enables the precise estimation of object positions within the CCTV's field of view as coordinates on the actual road network. In the proposed framework, these transformation results form the basis for constructing continuous trajectories, which are then utilized for path prediction and blind-spot inference in subsequent stages.

**Algorithm 2** Physics-based Perception (Agent 1)

**Require:** Pixel Seq $U$, Matrices $K, R$, Pos $C_w$, Step $\Delta t$
**Ensure:** State $\mathcal{S}_t = (\mathbf{p}_t, v_t, a_t, \theta_t)$

1: **procedure** COMPUTEPHYSICALSTATE($U, K, R, C_w$)
2: $\mathcal{P} \leftarrow []$
3: *// 1. Static Geometry (Ray Casting)*
4: **for** each $(u, v)$ in $U$ **do**
5: $\mathbf{d}_{cam} \leftarrow K^{-1}[u, v, 1]^T$
6: $\mathbf{d}_{world} \leftarrow R^{-1}\mathbf{d}_{cam}$
7: $\lambda \leftarrow -C_{w,z}/\mathbf{d}_{world,z}$
8: **if** $\lambda > 0$ **then**
9: $\mathbf{p}_{world} \leftarrow C_w + \lambda \mathbf{d}_{world}$
10: $\mathcal{P}$.append($\mathbf{p}_{world}$)
11: **else**
12: $\mathcal{P}$.append(NULL)
13: **end if**
14: **end for**
15: *// 2. Dynamic Kinematics*
16: $\mathbf{p}_{curr} \leftarrow \mathcal{P}[t]$, $\mathbf{p}_{prev} \leftarrow \mathcal{P}[t-1]$
17: $\vec{v} \leftarrow (\mathbf{p}_{curr} - \mathbf{p}_{prev})/\Delta t$
18: $v_t \leftarrow \|\vec{v}\|$ ▷ Speed
19: $v_{prev} \leftarrow \|(\mathbf{p}_{prev} - \mathcal{P}[t-2])/\Delta t\|$
20: $a_t \leftarrow (v_t - v_{prev})/\Delta t$ ▷ Accel
21: $\theta_t \leftarrow \text{atan2}(\vec{v}_y, \vec{v}_x)$ ▷ Heading
22: **return** $\mathcal{S}_t = (\mathbf{p}_{curr}, v_t, a_t, \theta_t)$
23: **end procedure**

Fig 4. Pseudo Algorithm for Agent 1

Fig 4 shows the procedural steps of Agent 1, which performs the pixel-to-world transformation and kinematic analysis as part of the **Semantic Orchestration** pipeline.

*Step 1: Static Geometry Projection (Lines 3–14)*

The core of this stage is **inverse projection** conducted through **ray casting**.

- **Line 5**: Pixel coordinates $(u, v)$ on the image plane are converted into a normalized direction vector $d_{cam}$ within the camera coordinate system using the inverse of the intrinsic parameter matrix $K^{-1}$.
- **Line 6**: This vector is then transformed into a direction vector $d_{world}$ relative to the world coordinate system by applying the inverse of the rotation matrix $R^{-1}$.
- **Line 7**: To identify the intersection point between the ray extending into world space and the ground plane ($z = 0$), the parameter $\lambda$ is calculated using the camera's height $C_{w,z}$ and the vertical component of the ray $d_{world,z}$.
- **Lines 8–10**: If the calculated $\lambda > 0$ (indicating the ray is directed toward the ground), the 3D world coordinates $p_{world}$ are derived and stored. Invalid rays, such as those directed toward the sky, are excluded from this process.

*Step 2: Dynamic Kinematics Analysis (Lines 15–21)*

Once the world coordinates are secured, the **Finite Difference Method** is utilized to derive kinematic characteristics.

- **Lines 16–17**: The velocity vector $\vec{v}$ is calculated by dividing the displacement between the current position $p_{curr}$ and the previous position $p_{prev}$ by the time interval $\Delta t$.
- **Line 18**: The scalar speed $v_t$ is determined through the magnitude of the velocity vector.
- **Lines 19–20**: Acceleration $a_t$ is derived from the rate of change in speed over time. This serves as a critical indicator for profiling driver behavior, such as aggressiveness.
- **Line 21**: Finally, the vehicle's heading angle $\theta_t$ is determined by applying the $atan2$ function to the components of the velocity vector.

The final estimated state $S_t = (p_{curr}, v_t, a_t, \theta_t)$ is then passed to **Agent 2** as the initial condition for path planning and further spatial reasoning.

### 3.3 Prediction of Movement Direction and Velocity through Vehicle Trajectory Analysis

To predict vehicle movement paths within blind spots, it is essential to quantitatively estimate the future heading and velocity of the vehicle based on its prior trajectory data. This section describes the methodology for analyzing movement patterns and classifying dynamic states—such as stopping, acceleration, and direction changes—using the vehicle trajectories previously converted into the world coordinate system, corresponding to step (7) in Fig. 2. First, the vehicle position data collected on a per-frame basis is chronologically sorted. Then, the displacement, velocity, and acceleration are calculated using the position differences between consecutive frames as follows:

$$d_t = \| \mathrm{p}_t - \mathrm{p}_{t-1} \|$$

$$v_t = \frac{d_t}{\Delta t}$$
$$a_t = \frac{v_t - v_{t-1}}{\Delta t}$$

where $p_t$ denotes the vehicle's position (world coordinates) at time $t$, and $\Delta\, t$ represents the time interval between frames. The movement direction is defined by the absolute heading angle $\theta_{abs}$. Specifically, $\theta_{abs}$ is the angle between the north direction and the direction vector formed by two consecutive vehicle positions. The relative heading angle is defined as $\theta_{rel}$ = $\theta_{abs}$ - $\theta_{cam}$, where $\theta_{cam}$ is the viewing direction of the corresponding CCTV in world coordinates. The calculated relative heading angle is categorized into an eight-direction system (F, FL, L, BL, B, BR, R, FR) to qualitatively represent the vehicle's driving characteristics.

The vehicle's dynamic state is classified according to the following threshold-based rules:

- **Stop**: $v_t < \epsilon_v$
- **Acceleration/Deceleration**: $|a_t| > \epsilon_a$
- **Sharp Turn**: $|\theta_{re}| > \epsilon_\theta$

Here, $\epsilon_v$, $\epsilon_a$, and $\epsilon_\theta$ are hyperparameters for velocity, acceleration, and rotation angle, respectively, used to adjust the sensitivity of the behavioral analysis. Additionally, by determining whether the calculated position and heading in each frame are contained within the CCTV's FOV polygon, the system dynamically estimates the vehicle's movement status and its proximity to the visibility boundaries. These derived parameters serve as critical inputs for the subsequent blind-spot path prediction and candidate CCTV selection processes.

### *3.4 Beam Search-based Blind-spot Path Prediction and CCTV Candidate Selection*

To maintain tracking continuity after a vehicle exits the field of view (FOV) of the last observed camera, this study proposes a method to predict movement paths within blind spots and identify subsequent candidate CCTVs. This process consists of four main stages: (i) estimating the vehicle's exit state, (ii) expanding paths on the road graph, (iii) calculating scores based on movement velocity and heading, and (iv) performing LLM-based refinement of CCTV candidate weights.

First, we construct a window of past frames based on the point in time the vehicle exits the FOV to summarize its average velocity and heading. To account for sudden noise or stop events, velocity and acceleration are estimated using a conservative median-based approach, while heading information is refined by accumulating movement vectors across frames. If the movement is negligible or the direction is ambiguous, the most recent valid heading or predefined orientation labels are used as alternative indicators. This summarized information defines the "exit state" of the vehicle immediately after it leaves the camera's coverage.

Subsequently, a beam search algorithm is applied to a road centerline-based graph to explore possible paths within the blind spot. At each expansion step, the algorithm dynamically evaluates candidate paths by considering their alignment with the vehicle's current trajectory, segment lengths, turning requirements, and the driver's profile. Paths aligning with a straight continuation are prioritized with potential acceleration, while sharp turns are penalized or pruned. The penalty for curvature is adjusted based on whether the driver is profiled as conservative or aggressive, allowing for flexible path prediction. The evaluation function generates a normalized score between 0 and 1 by synthesizing direction consistency, fitness with expected travel distance, turn intensity, and momentum maintenance.

At critical decision points such as intersections, the LLM intervenes to provide correction scores for candidate paths. The LLM receives road attributes—including connectivity, road ID changes, and heading alignment—in JSON format to determine the suitability of each candidate. To select the next likely CCTV on the optimized path, we utilize pre-calculated intersection areas between camera FOVs and the road centerline. If a predicted path segment intersects or is adjacent to a CCTV's FOV, that camera is identified as a candidate. Final weights for these candidates are determined based on entry angle, expected dwell time, and redundancy, with the LLM re-weighting the scores to derive the Top-k candidates.

This methodology integrates linguistic map documentation with LLM-based reasoning to enable real-time path prediction and CCTV inference without prior training, effectively maintaining tracking continuity in complex urban environments.

Fig 5 describes the proposed neuro-symbolic path planning and camera handoff prediction process. [2]The algorithm takes the initial physical state ($S_0$), the road network graph ($G$), and the driver's acceleration history ($\mathcal{H}_{acc}$) as inputs.

*Step 1: Driver Profiling (Line 2)*

Prior to path planning, the system estimates the driver's latent driving characteristics.

- **Line 2**: The **Aggressiveness Index ($Aggr$)** is calculated by identifying the 90th percentile of the acceleration history and normalizing it by the system's limit acceleration ( $a_{max}$ ). This parameter is used in subsequent stages to adjust the **curvature cost function** (e.g., the penalty for sharp turns is mitigated for aggressive drivers).

**Algorithm 3** Neuro-Symbolic Path Planning (Agent 2)

**Require:** State $\mathcal{S}_0$, Graph $G$, History $\mathcal{H}_{acc}$, Beam $B$

**Ensure:** Path $P^*$, Camera $C_{next}$

1: **procedure** PLANTRAJECTORY($\mathcal{S}_0, G, \mathcal{H}_{acc}$)
2: $Aggr \leftarrow$ Percentile($|\mathcal{H}_{acc}|$, 90)$/a_{max}$
3: $\mathcal{B}]\dashv\Updownarrow \leftarrow$ [{path : [$\mathcal{S}_0$], score : 0}]
4: **for** $k = 1$ **to** $K$ **do**
5: $\mathcal{C}\dashv\backslash\lceil\int \leftarrow$ []
6: **for** each $h$ in $\mathcal{B}]\dashv\Updownarrow$ **do**
7: $u \leftarrow h$.path[$-1$]
8: **for** each neighbor $v$ of $u$ **do**
9: $d_{pred} \leftarrow v_t\Delta t + 0.5a_{eff}\Delta t^2$
10: **if** Dist($u, v$) $> d_{pred} \times \eta$ **then continue**
11: **end if**
12: $S_{sym} \leftarrow w_d S_{dir} + w_s S_{spd} - S_{curv}(Aggr)$
13: $\Delta_{LLM} \leftarrow 0$
14: **if** IsInter($u$) **and** Branch $> 1$ **then**
15: $Prt \leftarrow$ BUILDPROMPT($Aggr, u, v$)
16: $P_{LLM} \leftarrow$ LLM($Prt$)
17: $\Delta_{LLM} \leftarrow \delta \cdot \log(\frac{\max(P_{LLM}, \epsilon)}{0.5})$
18: **end if**
19: $S_{new} \leftarrow h.$sc $+ S_{sym} + \Delta_{LLM}$
20: $\mathcal{C}\dashv\backslash\lceil\int$.add({$h$.path + [$v$], $S_{new}$})
21: **end for**
22: **end for**
23: $\mathcal{B}]\dashv\Updownarrow \leftarrow$ TOPK($\mathcal{C}\dashv\backslash\lceil\int, B$)
24: **end for**
25: $P^* \leftarrow \mathcal{B}]\dashv\Updownarrow$[0].path
26: $C_{next} \leftarrow$ NULL, $S_{max} \leftarrow -\infty$
27: **for** each Camera $C_k$ in Map **do**
28: $L_{ov} \leftarrow$ Len(Inter($P^*, C_k.\Omega_{fov}$))
29: $T_{dwell} \leftarrow L_{ov}/v_{avg}$
30: **if** $T_{dwell} \geq 0.5s$ **and** Score($C_k$) $> S_{max}$ **then**
31: $C_{next} \leftarrow C_k$, $S_{max} \leftarrow$ Score($C_k$)
32: **end if**
33: **end for**
34: **return** $P^*, C_{next}$
35: **end procedure**

Fig 5. Pseudo Algorithm for Agent 2

*Step 2: Probabilistic Beam Search (Lines 4–24)*

The core planning loop explores the most probable future paths using a beam search strategy with a beam width $B$.

- **Lines 9–11 (Physical Feasibility Verification)**: The physical reachability for each candidate neighbor node ($v$) is verified. If the actual map distance exceeds the kinematically predicted distance ( $d_{pred}$ ) by a predefined tolerance ($\eta$), the path is immediately pruned for violating physical consistency.

- **Line 12 (Heuristic Scoring)**: A **symbolic score ($S_{sym}$)** is calculated based on direction consistency ($S_{dir}$), speed matching ($S_{spd}$), and the curvature penalty ($S_{curv}$) adjusted by the $Aggr$ index.

*Step 3: Neuro-Symbolic Fusion (Lines 13–18)*

This stage represents the core novelty of this research, where the symbolic planner and the LLM's neural inference are integrated.

- **Line 14**: To ensure computational efficiency, the LLM is called selectively only at **intersections (ambiguous scenarios)** where multiple branches exist.

- **Lines 15–17**: A prompt containing the driver profile and geometric context is passed to the LLM. The probability returned by the LLM ($P_{LLM}$) is converted into a **Log-Odds term ($\Delta_{LLM}$)** and added to the existing path score. This allows semantic reasoning (e.g., identifying driving intent) to directly influence the numerical optimization process.

*Step 4: Simulation-based Handoff Prediction (Lines 26–31)*

Once the optimal path ($P^*$) is confirmed, the system selects the best camera to continue tracking.

- **Line 28**: Instead of simple distance matching, a simulation is performed to calculate the **geometric intersection length ( $L_{ov}$ )** between the predicted trajectory and each camera's field-of-view polygon ($\Omega_{fov}$).

- **Line 29**: The **dwell time ($T_{dwell}$)** is estimated by dividing the intersection length by the average velocity.

- **Lines 30–31**: The camera with the highest fitness score (e.g., shortest ETA) among candidates that satisfy the minimum dwell time condition ($\geq 0.5s$) is selected as the next handoff target ($C_{next}$).

## IV. EXPERIMENT

### *4.1 Simulation Environment Setup*

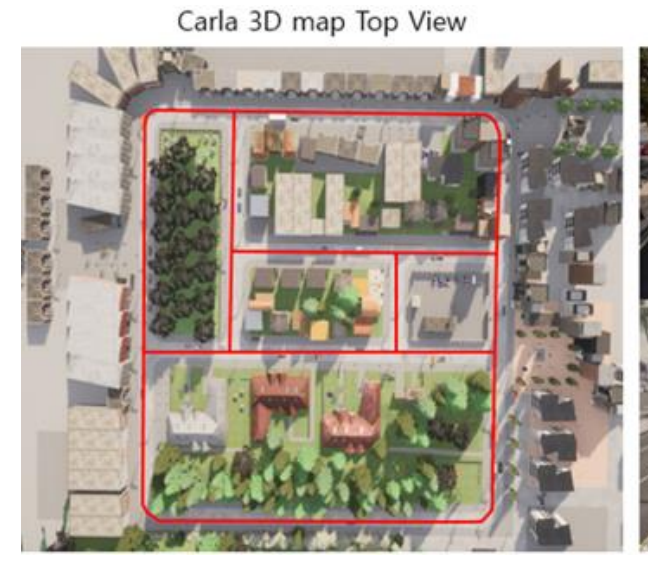

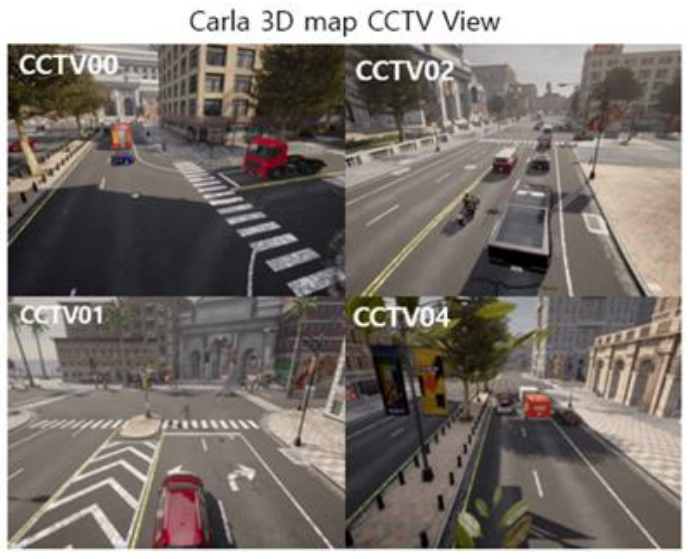

Fig 6. CARLA-based simulation environment for multi-CCTV data generation and system evaluation

The experimental environment was implemented using the **CARLA (0.9.15) simulator**, an open-source platform for autonomous driving research that provides high-fidelity urban physics and sensor models. The construction process is categorized into urban traffic network configuration, anomalous behavior generation, and the strategic deployment of the surveillance infrastructure.

#### *4.1.1 Urban Traffic and Vehicle Control*

We utilized a city-scale digital twin map (e.g., Town02) containing diverse intersection topologies and multi-lane road structures. To replicate a realistic urban traffic flow, approximately 30 autonomous vehicles were spawned and managed via the **CARLA Traffic Manager (TM)** module.

The TM module dynamically controlled vehicle speeds, lane-keeping, and intersection priorities based on the road network's centerline waypoints, which were extracted at 1-meter intervals to serve as the ground truth for our map-based documentation.

#### 4.1.2 Anomalous Trajectory Generation

To evaluate the robustness of the **SPOT** agent in non-ideal tracking scenarios, we implemented a custom behavioral agent to insert intermittent anomalous events. While the general traffic followed standard driving protocols, the target vehicle was programmed to perform:

**Sudden Kinematic Changes**: Intermittent rapid acceleration and abrupt braking to challenge the velocity-based prediction of Agent 1.

**Rapid Maneuvers**: Sudden lane changes and sharp turns at intersections to test the steering-habit profiling and beam search scoring of Agent 2.

#### 4.1.3 Surveillance Infrastructure and Sensor Configuration

CCTVs were strategically positioned at the summits of traffic light structures at major intersections to replicate actual urban surveillance visibility. The configuration process for each sensor unit involved:

**Placement & Extrinsics**: Each camera was assigned a unique ID and its 3D world coordinates $(x, y, z)$ and orientation (pitch, yaw, roll) were pre-cached to facilitate the **Ray-Plane intersection** analysis.

**Visual Constraints**: To simulate the most challenging multi-camera tracking conditions, CCTVs were arranged with **non-overlapping Fields of View (FOV)**, ensuring the existence of "blind spots" between monitoring zones.

**Sensor Parameters**: High-fidelity RGB camera sensors were configured with specific intrinsic parameters, including a 90-degree FOV and fixed focal lengths, to ensure parity between pixel-level detection and world-coordinate projection.

### 4.2 Data Configuration

To validate the performance of the proposed system, we generated various datasets required for object tracking and spatial reasoning using the CARLA simulator. The experimental data consists of four types: CCTV view images, CCTV world location documents, map waypoint documents, and structured documents for **Retrieval-Augmented Generation (RAG)**.

#### 4.2.1 CCTV View Image

This includes visual data input from the Field of View (FOV) of each CCTV placed in the simulated urban environment. Vehicles enter and exit the camera's view as they travel along the road network. These images serve as input for the object detection and tracking modules to generate pixel-coordinate-based trajectories per frame. CCTVs were arranged without visual overlap to construct tracking scenarios involving blind spots.

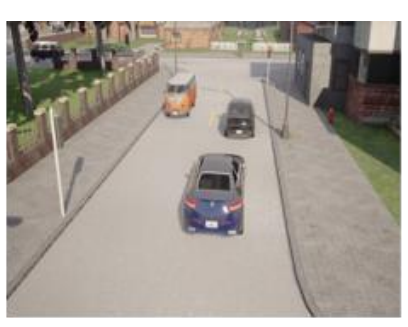
CCTV_05

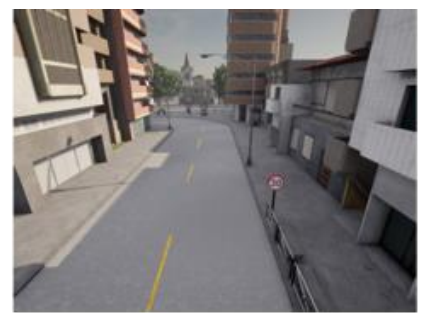
CCTV_07

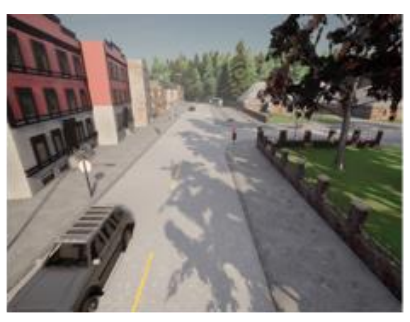
CCTV_03

Fig 7. CCTV view images

#### 4.2.2 CCTV World Location Document

This document structures the physical location information of each CCTV on the map, including its unique identifier (cctv_id), world coordinates $(x, y, z)$, installation height, and orientation. It is utilized as the extrinsic parameter for transforming pixel-level object positions to the world coordinate system (Section III-B) and as reference information for spatial relationships between FOV ranges and road structures.

#### 4.2.3 Map Waypoint Document

This data represents the urban road network using 2D coordinates. Waypoints extracted at regular intervals along road centerlines include the road identifier (road_id), lane identifier $(lane_id)$, intersection status $(is_intersection)$, and world coordinates $(x, y, z)$.

#### 4.2.4 Generated Document for RAG

These are high-level spatial documents created by combining map waypoint and CCTV location data. The documents are partitioned based on intersections; each document contains waypoint information for specific road segments, a list of monitoring CCTVs, FOV coverage ratios, and neighboring road segments. This structured set allows the LLM agent to utilize map space through RAG-based querying for path exploration and blind-spot inference.

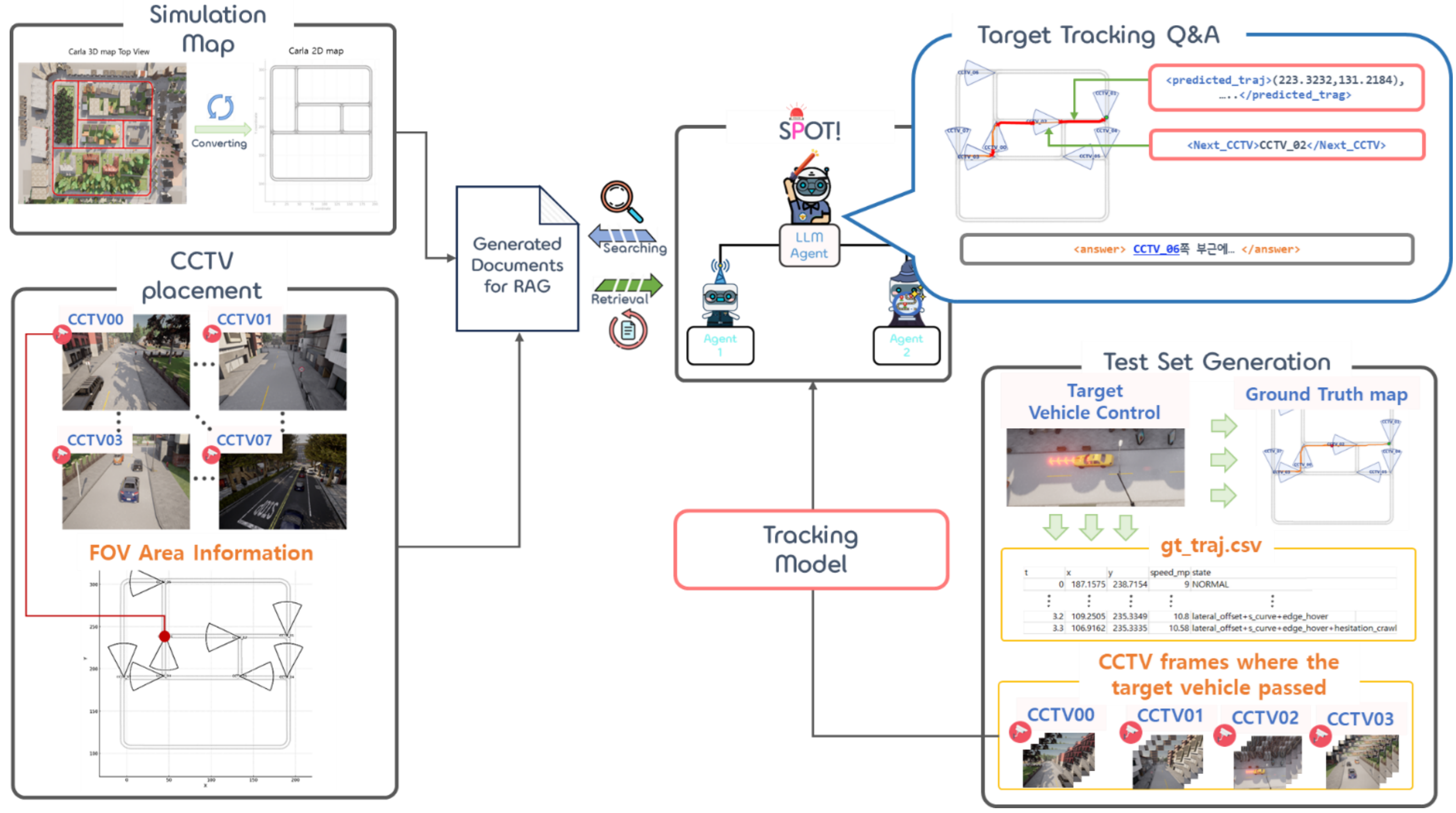

Fig 8. Evaluation Flow**:** (1) Discretization of the CARLA environment into 2D waypoints; (2) Strategic CCTV deployment (6m height, -20° pitch, 60° FOV) at intersections; (3) Construction of the RAG-based knowledge base; (4) Scenario-based ground-truth trajectory acquisition; and (5) Multi-agent orchestration (SPOT!) for behavioral analysis and blind-spot path planning.

### *4.3 System Specifications and Implementation Environment*

The experiments were performed on a hardware environment equipped with an Intel Core i7-13700K (3.40GHz) processor, 32GB RAM, and an NVIDIA GeForce RTX 3050 (8GB) GPU.

### *4.4 Performance Evaluation*

The evaluation metric assesses whether SPOT can predict a vehicle's movement path using only spatial knowledge derived from map documentation in a non-overlapping CCTV environment. We also verified the accuracy of identifying the next CCTV where the vehicle will appear. For this purpose, the actual next CCTV encountered after a vehicle exits an FOV was collected as the **Ground Truth (GT)**. Performance was measured by comparing SPOT's predicted candidates with the GT to analyze its ability to maintain tracking continuity across blind spots.

#### *4.4.1 Comparative Experimental Setup*

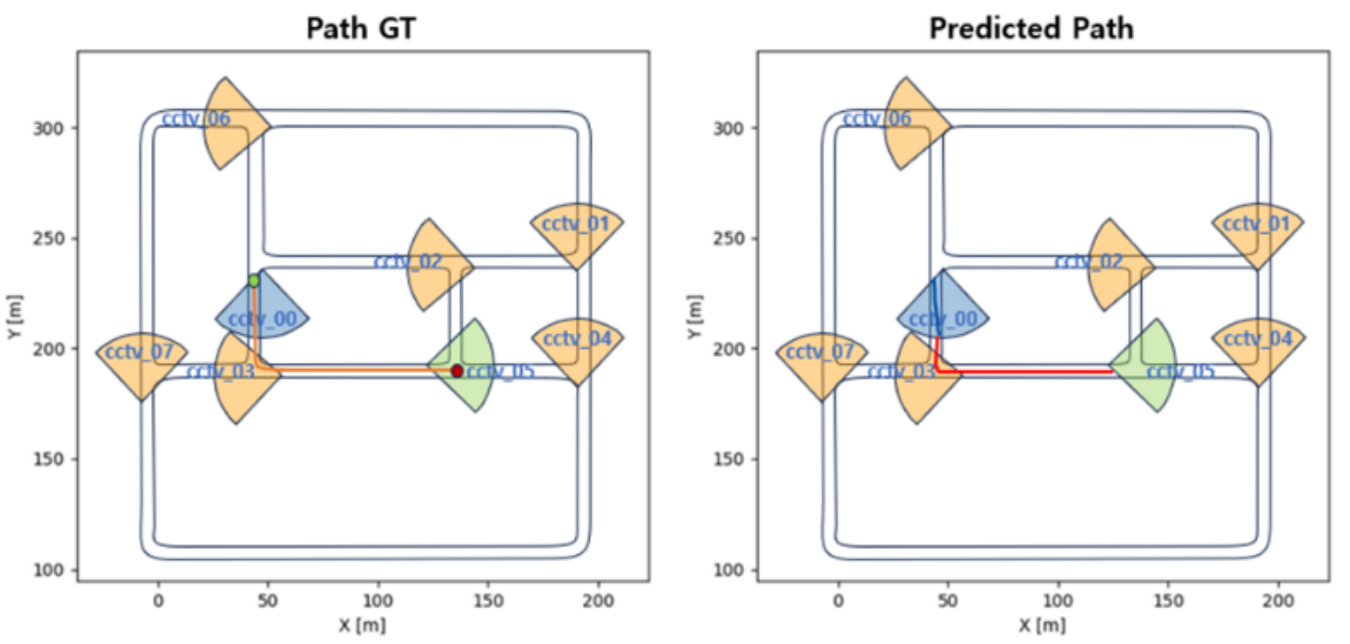

Fig 9. Visualization of the tracking handover process: the observed pixel trajectory is transformed into world coordinates (blue line), followed by blind-spot path planning to select the subsequent CCTV candidate (green FOV).

To evaluate the efficacy of the proposed SPOT framework, we conducted a comparative analysis across five distinct groups. These groups include four state-of-the-art Large Language Models (LLMs) and a purely heuristic baseline that operates without neural reasoning.

1. **DeepSeek-chat V3.2 (Neural-Symbolic Leader)**: Optimized for complex logical reasoning and code generation, DeepSeek-chat was selected due to its superior performance in handling structured data, such as the JSON-formatted waypoints and kinematic states used in our Agent 2 module.

2. **Llama-3-8B (Local Efficiency)**: As a representative of dense, instruction-tuned open-source models, Llama-3 provides a balance between reasoning depth and low-latency execution, making it a viable candidate for edge-side deployment in urban surveillance systems.

3. **GPT-4o-mini (Generalist Control)**: This model serves as a balanced benchmark, characterized by high stability in instruction following and consistent output formatting across diverse general-purpose tasks.

4. **Gemini-1.5-Flash (Long-Context Benchmark)**: Utilizing a Mixture-of-Experts (MoE) architecture with an extensive context window, this model was evaluated to determine the impact of long-range spatial information on path prediction.

5. **Heuristic Baseline (Deterministic Control)**: This group relies strictly on the _numeric_score function,

utilizing purely kinematic equations (e.g., direction consistency, velocity matching, and curvature penalties) without the neural-symbolic integration of an LLM.

#### *4.4.2 Performance Analysis*

Table I presents the quantitative evaluation of the proposed SPOT framework under different LLM configurations. Compared to the heuristic baseline without LLM reasoning, all LLM-based variants significantly reduce trajectory prediction errors and improve next-CCTV identification accuracy, demonstrating the effectiveness of map-based language reasoning in blind-spot scenarios.

Among the evaluated models, DeepSeek-chat 3.2 achieves the best overall performance, yielding the lowest ADE (3.64 m) and the highest Top-1 and Top-K CCTV accuracy (70% and 78%, respectively). These results indicate that integrating structured map documents with LLM-based spatial reasoning substantially enhances both trajectory prediction and next-camera estimation.

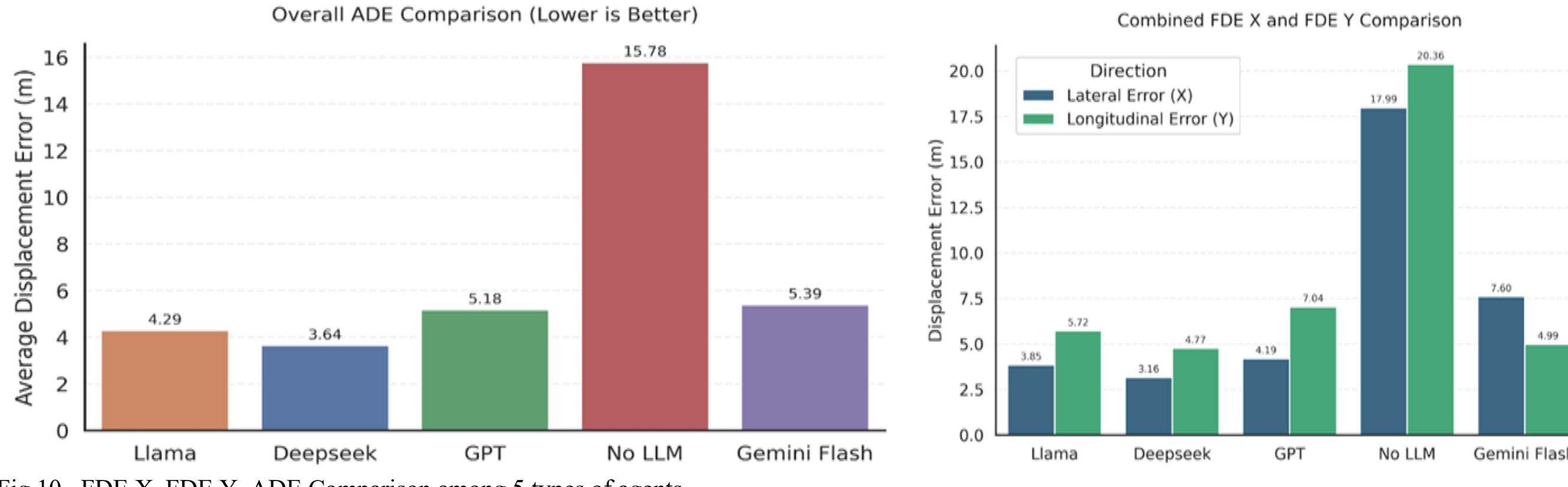

Fig 10. FDE X, FDE Y, ADE Comparison among 5 types of agents

TABLE 1. PERFORMANCE COMPARISON OF SPOT WITH DIFFERENT LLM CONFIGURATIONS FOR TRAJECTORY AND NEXT-CCTV PREDICTION

| Model Configuration | FDE(X) [m] | FDE(Y) [m] | ADE [m] | CCTV Top-1 (GT(T1)) | CCTV Top-K (GT(TK)) |
|---|---|---|---|---|---|
| + Heuristic (No LLM) | 9.35 | 9.84 | 15.78 | 27 | 28 |
| + GPT-4o-mini | 5.97 | 6.44 | 5.18 | 55 | 69 |
| + Llama-3-8B | 4.87 | 5.74 | 4.10 | 68 | 74 |
| + DeepSeek-chat 3.2 | **3.16** | **4.77** | **3.64** | **70** | **78** |
| + Gemini-flash 1.5 | 7.59 | 4.98 | 5.39 | 40 | 52 |

#### *4.4.3 Evaluation Metrics*

To rigorously assess the performance of the SPOT framework, we define evaluation metrics in two dimensions: Positional Precision and Systemic Handover Efficiency.

#### *1) Positional Error Reduction*

As illustrated in Fig 10, the Heuristic Baseline (Deterministic Control) suffered from significant error accumulation with an ADE of 15.78m. This confirms the limitation of relying strictly on numeric score functions without semantic reasoning. In contrast, DeepSeek-chat V3.2 (Neural-Symbolic Leader) achieved the highest spatial precision across all metrics, recording the lowest ADE (3.64m) and minimizing lateral error (FDE-X) to 3.16m. This superior performance aligns with its optimization for complex logical reasoning and structured data handling, proving its capability to accurately interpret the JSON-formatted kinematic states used in our Agent 2 module.

Gemini-1.5-Flash (Long-Context Benchmark) showed a unique performance profile. While its longitudinal error (FDE-Y) of 4.98m was competitive (2nd best), its lateral error (FDE-X) was the highest among LLMs at 7.59m. This suggests that while its extensive context window helps in maintaining the overall depth of the trajectory (long-range spatial awareness), it may lack the fine-grained local precision required for lateral adjustments compared to the Neural-Symbolic optimized DeepSeek.

#### *2) Handoff Success and Re-identification (Re-ID)*

Systemic success rates dramatically improved with neural reasoning. DeepSeek-chat led the field with 70 successful handoffs (Top-1), validating its status as the 'Neural-Symbolic Leader' in processing structured waypoints.
Notably, Llama-3.1-8B (Local Efficiency) demonstrated impressive robustness with 68 successful handoffs, closely trailing DeepSeek. Despite a slightly higher ADE (4.10m) than DeepSeek, its high success rate proves that it effectively balances reasoning depth with execution efficiency. This makes Llama-3.1 a highly viable candidate for edge-side deployment in urban surveillance, where maintaining systemic integrity is often more critical than achieving sub-meter coordinate perfection. GPT-4o-mini (Generalist Control) served as a stable benchmark with an ADE of 5.18m and 55 successful handoffs. While consistent, it fell short of the specialized performance shown by the Neural-Symbolic (DeepSeek) or Local Efficiency (Llama) models, highlighting the necessity of domain-specific optimization for complex trajectory prediction tasks.

## V. CONCLUSION

In this study, we proposed **SPOT (Spatial Prediction Over Trajectories)**, a map-guided LLM agent designed to link vehicle trajectories in an unsupervised manner across multi-CCTV environments with non-overlapping fields of view. The proposed framework structures road networks into waypoint-based documents and chunks them according to CCTV coverage, enabling the LLM to perform real-time spatial question-answering and path inference. By transforming pixel coordinates from CCTV images into the world coordinate system and employing a beam search algorithm that integrates vehicle heading and velocity, SPOT effectively infers candidate CCTVs where a vehicle is likely to reappear after traversing blind spots. Experimental results conducted in a CARLA-based urban simulation environment demonstrated that SPOT reliably estimates vehicle movement paths in blind spots without prior training and maintains consistent performance in identifying subsequent CCTVs. these findings suggest that the integration of map documentation and LLM-based spatial reasoning holds significant potential to complement or even replace existing learning-based multi-camera tracking techniques. Future research will focus on extending the tracking logic to ensure continuous vehicle ID association at the predicted subsequent CCTV locations. Additionally, we plan to validate the system's applicability in real-world urban CCTV environments, analyze its scalability in multi-vehicle and congested scenarios, and incorporate additional contextual metadata—such as traffic signals and real-time events—to further enhance the robustness of the spatial reasoning model.

## Core Contribution

- **Yujin Noh**

  Project Leader & Multi-object tracking

- **Inho Jake Park**

  CARLA simulator, LLM Agents & RAG algorithm

# APPENDIX

## List of contents

ⱡ: These authors contributed equally to this work. The authors are listed in alphabetical order by their last names.

## Appendix A: Orchestrator: Details of Semantic Orchestration & VLA System

The **Orchestrator** serves as the central engine that transforms the physical states from Agent 1 and the predicted trajectories from Agent 2 into a knowledge representation interpretable by the Large Language Model (LLM). It functions as a retrieval engine that identifies the optimal context for spatial reasoning.

### *A.1. VLA Tokenization (Visual-Language Alignment Tokenization)*

To bridge the **semantic gap** between continuous physics and discrete language, this study proposes **VLA Tokenization**. This technique transforms physical states and actions into standardized text templates.

#### *1) State Token ($T_s$)*

To embed the vehicle's current kinematic context, the system combines the camera ID, recent movement history ($H_{dir}$), and world coordinates ($P_{world}$):

$$T_{scene} = \text{"[SCENE] cam } = C_{id} \oplus \text{mv} = \vec{M}_{hist} \oplus pos = (x, y)\text{"}$$

Here, $H_{dir}$ represents a sequence of movement directions over the most recent $N$ frames (e.g., "F > FR > R"). This serves to project the vehicle's rotational inertia into a textual space for the LLM to process.

#### *2) Route Token ($T_r$)*

The route token expresses the **intent** of the candidate paths generated by Agent 2:

$$\mathrm{T_{route}} = \text{"}[ROUTE]\ to = C_{next} \oplus turn = Type \oplus sem = Tags\text{"}$$

#### *3) Document Token ($T_{doc}$)*

Static documents within the Retrieval-Augmented Generation (RAG) database, such as CCTV metadata and zone information, are structured into a searchable format:

$$\mathrm{T_{doc}} = \text{"}[DOC]\ type = ObjType \oplus id = ID \oplus sem = Attribute\text{"}$$

TABLE A.1. VLA Tokens Definition & Examples

| Token Type | Symbol | Template Format | Concrete Example |
|---|---|---|---|
| **State Token** | $T_{scene}$ | [SCENE] cam={ID} mv={Hist} pos={X,Y} | [SCENE] cam=C05 mv=F>FR>R pos=(120.5, 50.2) |
| **Route Token** | $T_{route}$ | [ROUTE] to={Next} turn={Type} sem={Tag} | [ROUTE] to=C07 turn=RIGHT sem=Intersection |
| **Knowledge Token** | $T_{doc}$ | [DOC] type={Type} id={ID} sem={Attr} | [DOC] type=Zone id=Z_02 sem=SchoolZone |

### *A.2. Semantic Alignment Mechanism*

By mapping the aforementioned tokens into a $d$-dimensional vector space $\mathbb{R}^d$ using a pre-trained embedding model $E(\cdot)$, the consistency between state and action can be judged solely through vector dot products, eliminating the need for explicit physical numerical values. This Triadic Relation is schematized as follows:

$$\underbrace{E(T_{scene})}_{\text{Query: Current State}} \quad \longleftrightarrow \quad \begin{cases} E(T_{route}) & \text{(Key: Future Action)} \\ E(T_{doc}) & \text{(Key: Static Knowledge)} \end{cases}$$

Operational Principle: If a vehicle is executing a right turn, the state vector $T_{scene}$ moves spatially closer to the right-turn route vector $T_{route}$ in the vector space, resulting in a high cosine similarity.

*A.3. Hybrid Retrieval & Reranking*

*1) Spatio-Semantic Filtering*

To ensure computational efficiency, an initial selection of candidates ($D_{cand}$) is performed with $O(1)$ complexity using R-Tree (for spatial data) and MinHash (for text deduplication).

$$\mathrm{S_{spatial}} = \{d \in \mathcal{D} \mid dist\big(d_{pos}, p_{curr}\big\} \leq R\}$$

$$\mathrm{S_{semantic}} = \{d \in \mathcal{D} \mid J\big(Q, d\big) \ \approx P\big(h_{min}(Q) = h_{min}(d)\big) \geq \tau\}$$

$$\mathrm{C_{cand}} = \mathrm{S_{spatial}} \cap S_{semantic}$$

*2) Alignment-based Reranking*

For the selected candidates $\mathrm{d_i}$ , the final ranking is determined by calculating the VLA alignment probability ($P_{align}$). The "probability of conforming to the current situation" is derived by passing the cosine similarity through a sigmoid function $\sigma$:

$$\mathrm{P_{align}}(key|query) = \sigma\left(E(T_{scene}) \cdot E\big(T_{key}\big)\right) = \frac{1}{1 + \exp\left(-E(T_{scene}) \cdot E\big(T_{key}\big)\right)}$$

The final score is derived as a weighted sum of the existing retrieval score $S_{base}$ and the alignment probability:

$$\mathrm{S_{final}} = (1 - \beta)S_{base} + \beta P_{align}$$

Here, the weight $\beta$ is dynamically adjusted based on the Context Richness (CR) of the retrieved information:

$$\mathrm{CR} = \min\left(1.0\,, \quad \frac{|C_{cand}|}{N_{optimal}}\right), \ \ \beta = \beta_{base} \times (0.3 + 0.7(1 - CR))$$

This mechanism is designed to rely more heavily on VLA-based reasoning in scenarios where information is scarce (low CR), thereby ensuring system robustness.

*A.4. Session Memory Management*

To maintain the context of vehicle movement trajectories during multi-turn conversations, a FIFO memory queue ($M$) with temporal decay is utilized. Whenever a user query ($Q$) is received, the most relevant historical trajectory information (Last Trajectory) is prioritized and injected into the context based on the following selection mechanism:

$$\mathrm{D_{context}} = TopK_\cos(E(Q), \{E(m) |\ m \in \mathcal{M}\}$$

In this process, the system computes the cosine similarity between the embedding of the current query $E(Q)$ and the embeddings of historical trajectories $E(m)$ stored in memory. By selecting the top-$K$ most relevant entries, the orchestrator ensures that the LLM's spatial reasoning is grounded in the most pertinent past movement patterns, even when navigating complex, multi-stage blind-spot scenarios.

*A.5. Map Comprehension Example*

[Response 1] Query: Please Explain about entire CCTVs on the map you found

### CCTV_00
- **Location**: (45.9, 238.2, 6.0)
- **Covered Roads**: 9
- **Total Coverage**: 261.2 m
- **Average Road Ratio**: 0.29
- **Neighboring Roads**: ['131', '8', '224', '92', '93', '118', '216', '232', '236']

### CCTV_01
- **Location**: (189.7, 239.4, 6.0)
- **Covered Roads**: 6
- **Total Coverage**: 267.7 m
- **Average Road Ratio**: 0.30
- **Neighboring Roads**: ['13', '31', '32', '47', '55', '62']

### CCTV_02
- **Location**: (134.0, 236.9, 6.0)
- **Covered Roads**: 6
- **Total Coverage**: 269.0 m
- **Average Road Ratio**: 0.26
- **Neighboring Roads**: ['134', '11', '142', '148', '177', '178']

### CCTV_03
- **Location**: (43.7, 191.6, 6.0)
- **Covered Roads**: 6
- **Total Coverage**: 266.5 m
- **Average Road Ratio**: 0.35
- **Neighboring Roads**: ['4', '199', '200', '224', '232', '236']

### CCTV_04
- **Location**: (189.7, 190.1, 6.0)
- **Covered Roads**: 6
- **Total Coverage**: 268.9 m
- **Average Road Ratio**: 0.36
- **Neighboring Roads**: ['277', '14', '276', '283', '252', '290']

### CCTV_05
- **Location**: (133.8, 191.3, 6.0)
- **Covered Roads**: 6
- **Total Coverage**: 257.5 m
- **Average Road Ratio**: 0.34
- **Neighboring Roads**: ['6', '317', '308', '322', '341', '342']

### CCTV_06
- **Location**: (43.7, 302.5, 6.0)
- **Covered Roads**: 7
- **Total Coverage**: 273.2 m
- **Average Road Ratio**: 0.32
- **Neighboring Roads**: ['2', '18', '351', '383', '384', '385', '390']

### CCTV_07
- **Location**: (-3.5, 190.2, 6.0)
- **Covered Roads**: 6
- **Total Coverage**: 267.7 m
- **Average Road Ratio**: 0.26
- **Neighboring Roads**: ['0', '411', '412', '426', '433', '437']

Fig A.1. In Response 1 investigated the LLM's comprehension of holistic map information, specifically querying about CCTV placement and road coverage status.

[Response 2] Query: Please fully describe the map you have

=== Answer ===
The map structure consists of a comprehensive network of waypoints, zones, and CCTVs designed for effective surveillance and monitoring. Here's a detailed overview:

### Map Structure Overview

1. **Waypoints**:
   - The map contains a total of **2,972 waypoints**. These waypoints serve as reference points for navigation and tracking within the mapped area.

2. **Zones**:
   - There are **13 refined zones** on the map. Each zone is defined by specific roads and is associated with one or more CCTVs. The zones help in organizing the surveillance coverage and managing the areas effectively.

3. **CCTVs**:
   - The map features **seven CCTVs**: CCTV_00, CCTV_01, CCTV_02, CCTV_03, CCTV_04, CCTV_05, CCTV_06, and CCTV_07. Each CCTV is strategically placed to cover specific roads and zones, enhancing the overall surveillance capability.

### Zone Details
- Each zone has a unique identifier and is characterized by:
  - **Covered Roads**: A list of roads that fall within the zone.
  - **CCTVs**: The specific CCTVs that monitor the zone.
  - **Neighbors (Zones)**: Information about adjacent zones, which helps in understanding the connectivity and coverage.
  - **Average Exposure Ratio**: This metric indicates the effectiveness of the CCTV coverage in the zone.

### CCTV Details
- Each CCTV has specific attributes:
  - **Location**: The geographical coordinates of the CCTV.
  - **Covered Roads**: The number of roads that the CCTV can monitor.
  - **Total Coverage**: The total distance covered by the CCTV in meters.
  - **Average Road Ratio**: A measure of how effectively the CCTV covers the roads in its vicinity.
  - **Neighbors (Roads)**: A list of roads adjacent to the CCTV's coverage area.

### Example Zones and CCTVs
- **Zone Z_-1_2**: Monitored by CCTV_07, covering roads like '0', '1', '2', etc. It has no neighboring zones and an average exposure ratio of 0.23.
- **Zone Z_0_3**: Monitored by CCTV_06, covering roads such as '384', '385', and others, also with no neighboring zones and an average exposure ratio of 0.32.
- **CCTV_00**: Located at (45.9, 238.2), covering 9 roads with a total coverage of 261.2 m and an average road ratio of 0.29.

### Conclusion
The map is structured to provide comprehensive surveillance through a well-defined network of zones and CCTVs, ensuring that all critical areas are monitored effectively. The organization of waypoints and the strategic placement of CCTVs enhance the operational efficiency of the surveillance system.

Fig A.2. Response 2 is designed to evaluate the LLM's holistic comprehension of the entire simulation environment rather than its perception of a single CCTV node. Through this output, the LLM explicitly articulates the diverse informational layers embedded within the grid map—including waypoints, functional zones, and CCTV coordinates—thereby demonstrating its capability to maintain high-level spatial awareness across the global coordinate system.

[Response 3] Query: Please Explain about Zone that you can see

Here's a detailed explanation of the zones present on the map:

### Zone Overview
The map consists of **13 refined zones**, each designed to optimize surveillance and monitoring through strategically placed CCTVs. Each zone is characterized by its unique identifier, coverage area, and associated CCTVs.

### Zone Details

1. **Zone Z_0_1**
   - **CCTVs**: CCTV_00
   - **Covered Roads**: Includes major roads like Road 8, which has 100% coverage by CCTV_00.
   - **Average Exposure Ratio**: High, due to extensive coverage.

2. **Zone Z_0_2**
   - **CCTVs**: CCTV_01
   - **Covered Roads**: Includes roads such as Road 216 and Road 199.
   - **Average Exposure Ratio**: Moderate, with effective monitoring of key intersections.

3. **Zone Z_0_3**
   - **CCTVs**: CCTV_02
   - **Covered Roads**: Includes Road 177, which has significant coverage.
   - **Average Exposure Ratio**: Good, ensuring critical areas are monitored.

4. **Zone Z_0_4**
   - **CCTVs**: CCTV_03
   - **Covered Roads**: Primarily covers Road 199, providing substantial surveillance.
   - **Average Exposure Ratio**: High, focusing on high-traffic areas.

5. **Zone Z_0_5**
   - **CCTVs**: CCTV_04
   - **Covered Roads**: Covers roads like Road 290 and Road 283.
   - **Average Exposure Ratio**: Moderate, with a focus on specific zones.

6. **Zone Z_0_6**
   - **CCTVs**: CCTV_05
   - **Covered Roads**: Includes several roads with moderate coverage.
   - **Average Exposure Ratio**: Balanced, ensuring adequate monitoring.

7. **Zone Z_0_7**
   - **CCTVs**: CCTV_06
   - **Covered Roads**: Covers Road 384 and others, ensuring critical areas are monitored.
   - **Average Exposure Ratio**: High, focusing on high-traffic zones.

8. **Zone Z_0_8**
   - **CCTVs**: CCTV_07
   - **Covered Roads**: Includes Road 412, providing substantial coverage.
   - **Average Exposure Ratio**: Good, ensuring effective monitoring.

9. **Zone Z_0_9**
   - **CCTVs**: CCTV_00
   - **Covered Roads**: Overlaps with Zone Z_0_1, enhancing coverage.
   - **Average Exposure Ratio**: Very high, due to redundancy in monitoring.

10. **Zone Z_0_10**
    - **CCTVs**: CCTV_01
    - **Covered Roads**: Similar to Zone Z_0_2, ensuring comprehensive coverage.
    - **Average Exposure Ratio**: High, focusing on critical intersections.

11. **Zone Z_0_11**
    - **CCTVs**: CCTV_02
    - **Covered Roads**: Similar to Zone Z_0_3, ensuring comprehensive coverage.
    - **Average Exposure Ratio**: High, focusing on critical intersections.

12. **Zone Z_0_12**
    - **CCTVs**: CCTV_03
    - **Covered Roads**: Similar to Zone Z_0_4, ensuring comprehensive coverage.
    - **Average Exposure Ratio**: High, focusing on critical intersections.

13. **Zone Z_0_13**
    - **CCTVs**: CCTV_04
    - **Covered Roads**: Similar to Zone Z_0_5, ensuring comprehensive coverage.
    - **Average Exposure Ratio**: Moderate, focusing on specific zones.

### Conclusion
Each zone is designed to maximize the effectiveness of the CCTV network, ensuring comprehensive coverage of critical areas while maintaining a balance between monitoring and resource allocation. The strategic placement of CCTVs within these zones enhances the overall surveillance capability of the map.

Fig A.3 Response 3 provides an in-depth investigation into the "Zone" abstraction, specifically evaluating the LLM's comprehension of spatial hierarchies derived from R-tree indexing. This inquiry focuses on how the model interprets localized spatial data within a defined zone and examines its capacity to discern relational dependencies between various environmental entities.

[Response 4] Query: What are the neighboring roads for Zone Z_0_3?

The neighboring roads for **Zone Z_0_3** are as follows:

- Road 4
- Road 5
- Road 216
- Road 224
- Road 232
- Road 236

These roads are adjacent to Zone Z_0_3, providing context for traffic flow and potential monitoring needs in the area.

Fig A.4. Response 4 presents a query designed to evaluate the LLM's foundational knowledge and factual accuracy regarding zone-specific information. This stage focuses on verifying whether the model correctly maintains the attributes and spatial constraints associated with each defined zone.

[Response 5] Query: From CCTV_04 to CCTV_07, please tell me the shortest route

=== Answer ===
Here is the shortest route from CCTV_04 to CCTV_07:

1. **Start at CCTV_04**: located at (189.7, 190.1).
2. **Identify the Route**:
   - Move from CCTV_04 along **Road 14** towards **Road 426**.
   - Continue straight on **Road 426** until reaching the intersection with **Road 0**.
   - Finally, proceed to CCTV_07 located at (-3.5, 190.2).

3. **End at CCTV_07**: located at (-3.5, 190.2).

### Waypoints along the route
1. (189.7, 190.1) - CCTV_04
2. (189.0, 190.1) - wp_001
3. (188.0, 190.1) - wp_002
4. (187.0, 190.1) - wp_003
5. (186.0, 190.1) - wp_004
6. (185.0, 190.1) - wp_005
7. (184.0, 190.1) - wp_006
8. (183.0, 190.1) - wp_007
9. (182.0, 190.1) - wp_008
10. (181.0, 190.1) - wp_009
11. (180.0, 190.1) - wp_010
12. (179.0, 190.1) - wp_011
13. (178.0, 190.1) - wp_012
14. (177.0, 190.1) - wp_013
15. (176.0, 190.1) - wp_014
16. (175.0, 190.1) - wp_015
17. (174.0, 190.1) - wp_016
18. (173.0, 190.1) - wp_017
19. (172.0, 190.1) - wp_018
...
181. (10.0, 190.1) - wp_180
182. (9.0, 190.1) - wp_181
183. (8.0, 190.1) - wp_182
184. (7.0, 190.1) - wp_183
185. (6.0, 190.1) - wp_184
186. (5.0, 190.1) - wp_185
187. (4.0, 190.1) - wp_186
188. (3.0, 190.1) - wp_187
189. (2.0, 190.1) - wp_188
190. (1.0, 190.1) - wp_189
191. (0.0, 190.1) - wp_190
192. (-1.0, 190.1) - wp_191
193. (-2.0, 190.1) - wp_192
194. (-3.0, 190.1) - wp_193
195. (-3.5, 190.2) - CCTV_07

This route provides a direct path to reach CCTV_07 quickly. Please proceed with caution and monitor the situation closely.

Fig A.5. Response 5 aims to leverage the LLM's internalized spatial representation to analyze the relative positioning and geographical distribution between discrete CCTV nodes. The primary objective is to facilitate consistent trajectory tracking by establishing a topological link between cameras, ensuring the system can navigate and follow paths across the broader simulation environment.

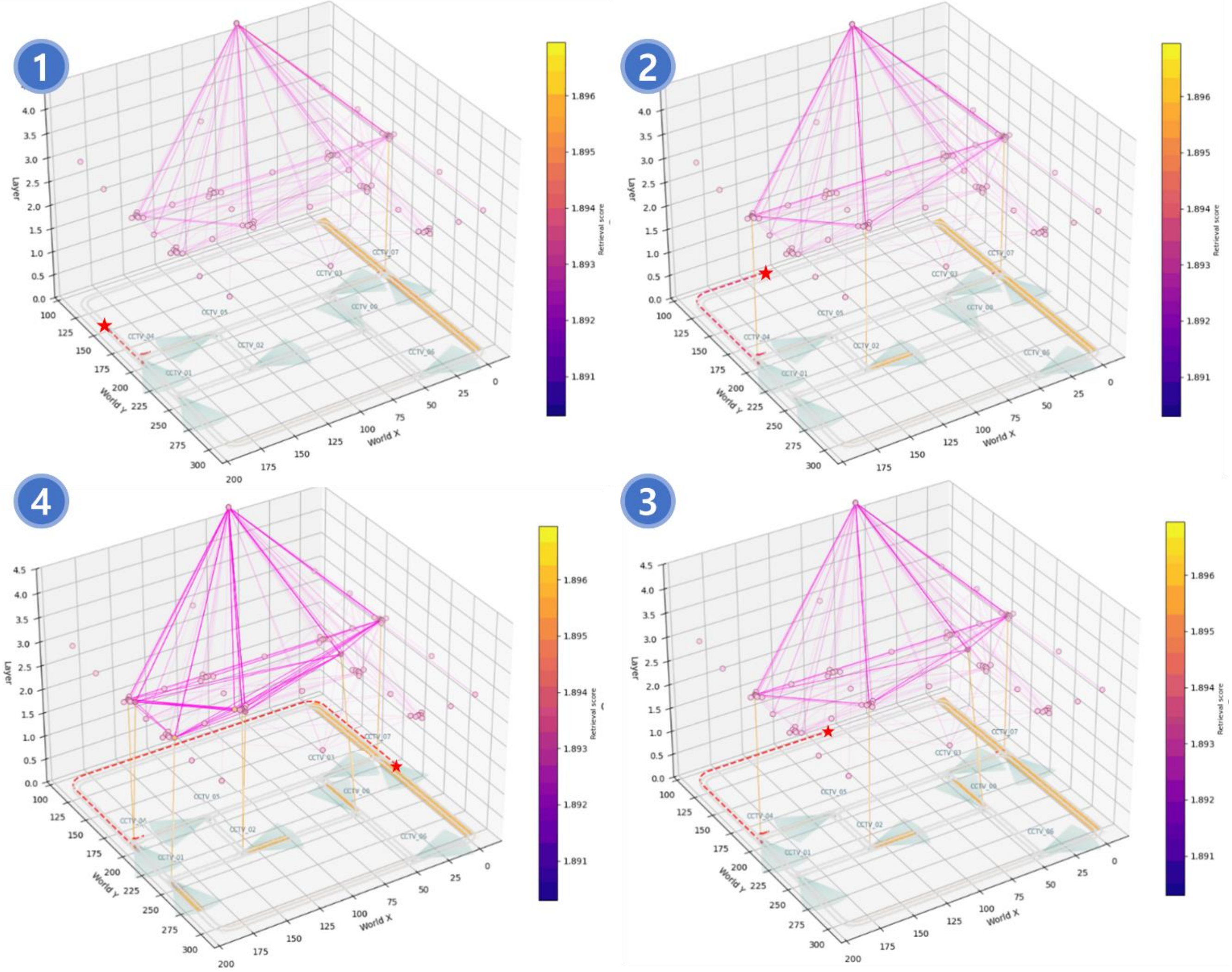

Fig A.6. The sequence progresses linearly from Stage 1 through Stage 4, during which the LLM formulates the optimal trajectory between CCTV_04 and CCTV_07. By analyzing the spatial topology, the model identifies and generates what it determines to be the shortest path connecting these two surveillance nodes, demonstrating its capacity for efficient inter-node navigation.

## APPENDIX B : AGENT 1: DETAILS OF PHYSICS-BASED PERCEPTION

The perception agent (Agent 1) functions as a deterministic physics engine that transforms 2D pixel coordinates from surveillance cameras into 3D world coordinate physical states, including position, velocity, and acceleration. This process is divided into two primary stages: **Static Geometric Projection** and **Dynamic Kinematic Analysis**.*A.1) Inverse Pinhole & Ray-Ground Intersection*

### *B.1. Inverse Pinhole & Ray-Ground Intersection*

To recover 3D world coordinates $P_{world} = [x, y, z]^T$ from pixel coordinates $u = [u, v, 1]^T$, this system utilizes an inverse pinhole camera model combined with ray-casting techniques, assuming a flat ground plane ($Z = 0$).

#### 1. Intrinsic Unprojection

Pixel coordinates $u$ are normalized into a camera coordinate ray vector $d_{cam}$ using the intrinsic parameter matrix $K$:

$$\mathrm{K} = \begin{bmatrix} f_x & 0 & c_x \\ 0 & f_y & c_y \\ 0 & 0 & 1 \end{bmatrix}, \tilde{u} = \begin{bmatrix} u \\ v \\ 1 \end{bmatrix}$$

The normalized vector is calculated as $d_{cam} = K^{-1}$u.

#### 2. Extrinsic Transformation

The ray vector is converted to the world coordinate system using a rotation matrix $R$. To ensure compatibility between CARLA (left-handed) and OpenCV (right-handed) systems, a coordinate transformation matrix $T_{coord}$ is applied:

$$\mathrm{d_{world}} = R^{-1}(T_{coord} \cdot d_{cam})$$

The ray $p(t)$ starting from the camera's world position $C_w$ is parameterized as follow.

$$\mathrm{p(t)} = C_w + t \cdot \frac{d_{world}}{||d_{world}||}, (t > 0)$$

**Ground Mapping**: Rays pointing toward the sky ($d_{world,z} \geq 0$) or originating behind the camera ($t < 0$) are discarded. For valid rays, the final world coordinate $p_{ground}$ on the road surface is derived:

$$\mathrm{p_{ground}} = \begin{bmatrix} C_{w,x} + t \cdot d_{world,x} \\ C_{w,y} + t \cdot d_{world,y} \\ 0 \end{bmatrix}$$

*B.2. Kinematic State Estimation*

The sequence of projected world coordinates $P = \{p_0, p_1, \ldots, p_t\}$ is used to estimate the vehicle's kinematic state $S_t = (v_t, a_t, \theta_t)$.

- **Velocity and Acceleration**

  To mitigate sensor noise, the finite difference method is applied over a time interval $\Delta t$.

  - **Velocity**: $v_t = \frac{p_t - p_{t-1}}{\Delta t}$, and scalar speed is $v_t = |v_t|$.
  - **Acceleration**: $a_t = \frac{v_t - v_{t-1}}{\Delta t}$, this serves as a key metric for profiling driver behavior (aggressive vs. defensive).

- **Heading and Turn Detection**

  - The absolute heading angle is calculated as $\theta_{world} = \mathrm{atan2}(\Delta y, \Delta x)$.
  - The relative azimuth $\theta_{rel} = (\theta_{world} - \psi_{cam})\ (mod 2\pi)$ is computed relative to the camera's line of sight $\psi_{cam}$ .
  - If the angular change $\Delta \theta_{rel}$ exceeds a threshold $\delta_\theta$, it is classified as a "Turn Event," which adjusts direction weights during the beam search.

$$\mathrm{Turn_t} = \begin{cases} Left & if\ \Delta\theta_{rel} > \delta_\theta \\ Right & if\ \Delta\theta_{rel} > -\delta_\theta \\ Straight & otherwise \end{cases}$$

  The physical state $S_t$ derived through this process is passed to **Agent 2**'s path planning algorithm as the initial state value.

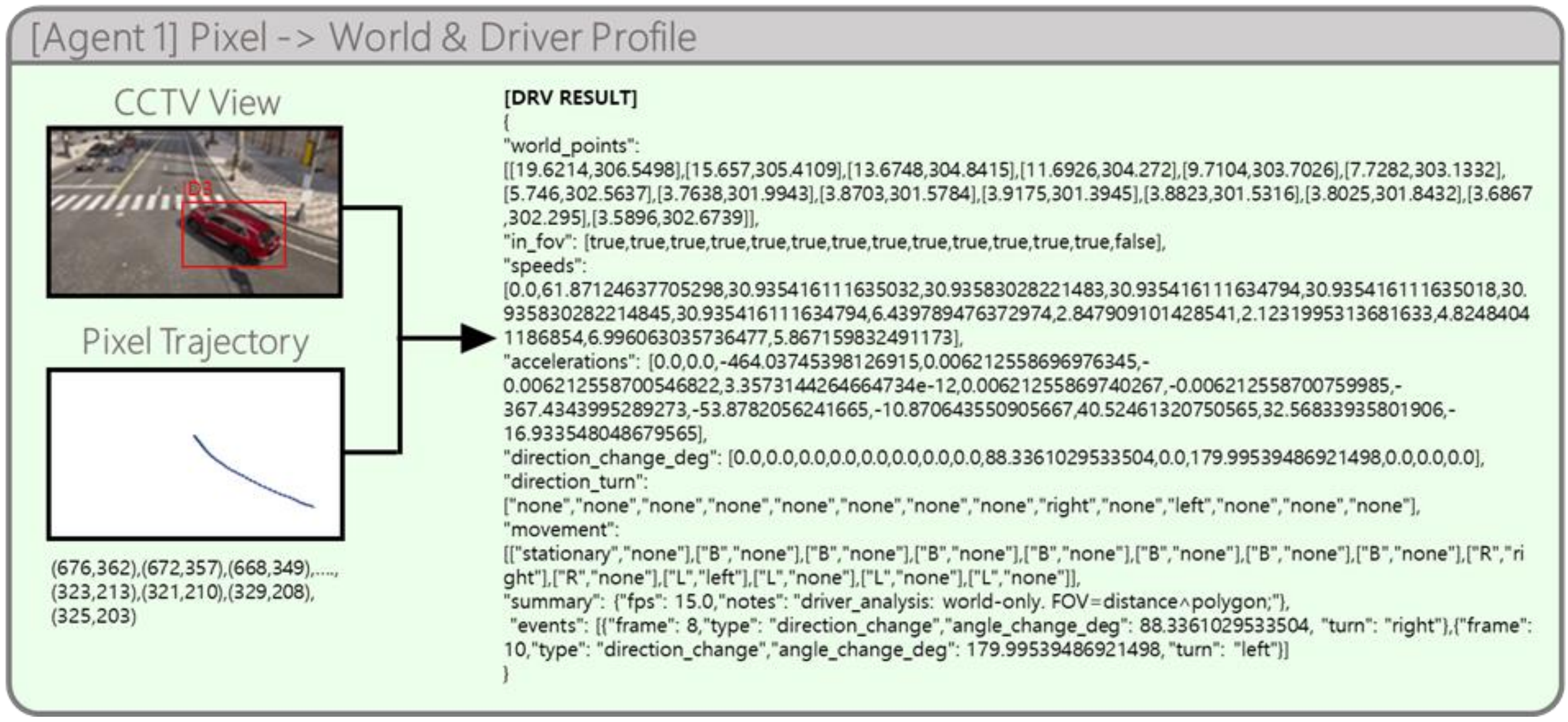

Fig B.1. This example illustrates the process of mapping pixel-level vehicle trajectories detected by CCTV into a world coordinate system, while concurrently performing a comprehensive analysis of the driver's behavioral profile. By transforming image-plane coordinates into global world-space data, the system enables the LLM to interpret both the physical movement and the underlying intent of the target vehicle in parallel.

## APPENDIX C : AGENT 2: DETAILS OF NEURO-SYMBOLIC PATH PLANNING

The Planning Agent (Agent 2) takes the physical state generated by Agent 1 as the initial value to probabilistically predict future vehicle trajectories and determine the optimal CCTV handoff timing. This process consists of Driver Profiling, Probabilistic Beam Search, Neuro-Symbolic Fusion, and Dynamic Handoff Simulation.

### *C.1. Driver Profiling*

The driver's aggressiveness is quantified using past observation data $\mathcal{H}_{\mathrm{acc}} = \{|\alpha_{t-N}|, \dots, |\alpha_t|\}$. Instead of the mean value, the **90th percentile** is used to capture the tendency of sudden acceleration and braking:

$$Aggr = \min(1.0\,, \frac{Percentile(\mathcal{H}_{acc}, 90)}{a_{cap}})$$

Here, $a_{cap}$ is the system-defined limit acceleration (e.g., $4.0\ m/s^2$). The calculated value $Aggr \in [0, 1]$ is used to dynamically adjust the **curvature penalty sensitivity ($\alpha$)** and the **search radius ($\eta$)** during subsequent path exploration.

### *C.2. Probabilistic Beam Search Cost Function*

When expanding paths on the graph via beam search ($B = 8$), the transition cost $S(u, v)$ from the current node $u$ to a neighbor node $v$ is calculated by synthesizing physical consistency and rule compliance as follows:

$$S_{total} = w_{dir}S_{dir} + w_{spd}S_{spd} + w_{rule}S_{rule} - S_{curv} + w_{mom}S_{mom}$$

The definition of each term is as follows:

1. **Direction Consistency ($S_{dir}$)**: This evaluates the alignment between the vehicle's inertial direction and the road by normalizing the cosine between the observed movement vector and the road link vector.

$$\mathrm{S_{dir}} = 0.5\left(\frac{v_{obs} \cdot e_{uv}}{||v_{obs}||\,||e_{uv}||}\right) + 1$$

2. **Kinematic Consistency ($S_{spd}$)**: The relative error $\epsilon$ between the predicted arrival distance $d_{pred}$, calculated with current physical quantities, and the actual map distance $d_{map}$ is evaluated using a **Gaussian Kernel**.

$$\mathrm{d_{pred}} = v_t \Delta t + \frac{1}{2} a_{eff} \Delta t^2, \epsilon = \frac{|d_{map} - d_{pred}|}{d_{pred}}$$

$$\mathrm{S_{spd}} = \exp\left(-\left(\frac{\epsilon}{\sigma}\right)^2\right)$$

The Gaussian kernel assigns a probabilistic score to the relative error $\epsilon$, exponentially suppressing paths that violate physical laws, such as nodes that are too far or too close.

**Dynamic Curvature Penalty ($S_{curv}$)**: While this is a deduction factor for the rotation angle $\theta$, the penalty is weighted by the physical fact that higher current velocity $v_t$ makes turning more difficult due to centrifugal force. $v_{ref}$ is set to the average velocity in typical urban driving.

$$\mathrm{S_{curv}} = \lambda \cdot \theta \cdot (1 + \gamma \frac{v_t}{v_{ref}})$$

*C.3. Neuro-Symbolic Log-Odds Fusion*

In sections where heuristic cost functions alone provide ambiguous results, such as complex intersections, the system invokes a Large Language Model (LLM) to perform semantic reasoning. This process is not a strict Bayesian inference but rather a method of injecting the LLM's confidence into the numerical search space in the form of Log-Odds.

*1) Prompt Engineering*

The LLM is assigned the persona of a **"Strategic Navigation Supervisor"** rather than being treated as a simple data processor. It is provided with a structured prompt as follow

TABLE C.1. LLM PROMPT STRUCTURE FOR BEAM SEARCH BRANCH VALIDATION

| Category | Content |
|---|---|
| System Role | "You are a Strategic Navigation Supervisor. Your role is to validate the heuristic path suggestion by analyzing the **Driver's Profile** and **Road Context**." |
| Input Context | 1. **Driver Profile: Aggr** (0.0~1.0) & Turn Intent (e.g., PREP_LEFT)<br>2. **Current State:** Speed ($v_t$), Location (u)<br>3. **Branches:** List of candidates $\{v_1, v_2, \dots\}$ with attributes:<br>- Turn Angle ( $\theta_{deg}$ )<br>- Semantic Tags (e.g., Intersection, LeftOnly)<br>- Physical Feasibility Check |
| Reasoning Logic | 1. **Consistency:** Does the branch match the driver's intent?<br>2. **Feasibility:** Is the turn physically possible at current speed?<br>3. **Safety:** Does it violate semantic lane rules? |
| Output Format | JSON: { "branches": [ {"id": v_i, "score": 0.0~1.0, "reason": "..."} ] } |

*2) Score Fusion*

To integrate the probability value returned by the LLM with the existing numerical score system (which operates in Log-Space), the following transformation is applied:

$$\Delta\mathrm{LLM} = \delta \cdot \log\left(\frac{\max(P_{LLM}, \epsilon)}{1 - \max(P_{LLM}, \epsilon)}\right) \approx \delta \cdot \log\left(\frac{P_{LLM}}{0.5}\right)$$

$$\mathrm{S_{final}}(u, v) = S_{heuristic}(u, v) + \Delta LLM$$

- **Neutral Point**: When $P_{LLM} = 0.5$, then $\Delta_{LLM} = 0$, meaning the system relies solely on the physical/heuristic scores.

- **Boosting & Suppression**: If $P_{LLM} > 0.5$, the path score is **boosted**; if $P_{LLM} < 0.5$, the score is **suppressed**. This mechanism serves to correct the limitations of numerical calculations through common-sense semantic reasoning.

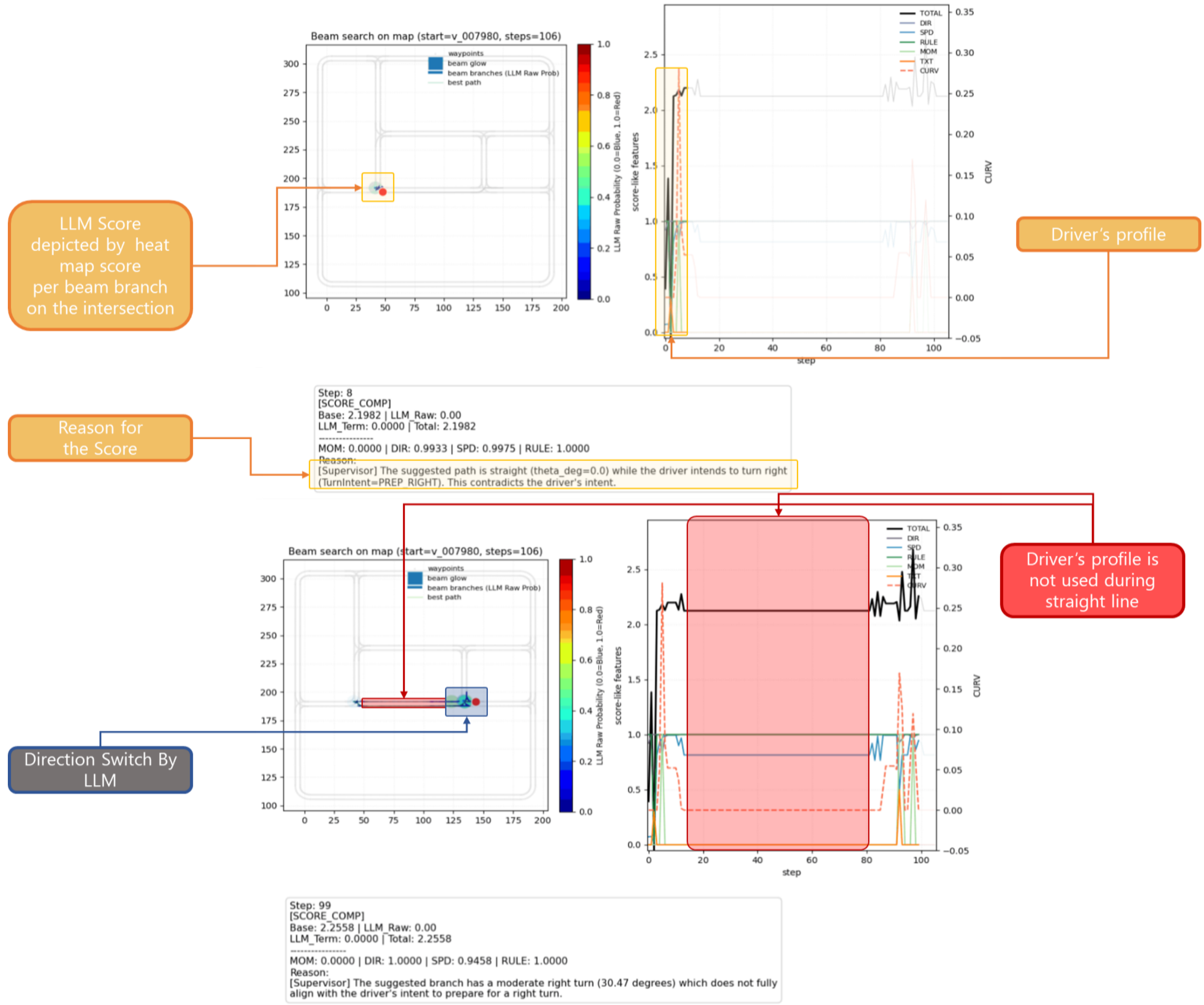

Fig C.1. This example demonstrates the LLM-driven decision-making process at an intersection. The system synthesizes multiple data streams—including driver profiles, observed historical trajectories, and structural intersection metadata—to evaluate candidate branches generated through Beam Search. By assigning quantitative scores to each branch based on these integrated inputs, the orchestrator executes blind-spot path planning, effectively predicting the vehicle's movement within occluded or unmonitored regions.

### *C.4. Simulation-based Handoff*

For the finalized optimal path $P^*$, a Spatio-Temporal Simulation is performed to determine the next tracking handoff. In this stage, the LLM is once again utilized to evaluate "ease of monitoring"—a qualitative metric that is difficult to determine through geometric values alone.

#### *1) Dynamic Dwell Time Integral*

The dwell time is integrated using the intersection length between each path segment $L_i$ and the field-of-view (FOV) polygon $\Omega_k$ of CCTV $k$.

$$T_{\text{dwell}}(k) = \sum_{i \in P} \frac{Length(L_i \cap \Omega_k)}{v_i}$$

TABLE C.2. LLM PROMPT STRUCTURE FOR CCTV-TOP K AND CCTV-TOP 1

| Category | Content |
| --- | --- |
| System Role | "You are an assistant scoring CCTV candidates for optimal vehicle tracking." |
| Scoring Criteria | 1. **Dwell Time:** High priority if ≥ 1.0s.<br>2. **Approach Angle:** Penalize 90-degree (perpendicular) crossings; prefer 30-60 degrees.<br>3. **ETA:** Shorter is better.<br>4. **Speed:** High speed targets require wider FOV coverage. |
| **Input Data** | JSON list of candidates with: { "id", "eta_s", "dwell_s", "angle_deg", "speed_m_s" } |

[Agent 2] Beam Search

Beam Branch value on the road

```
{
  "llm_reason": "",
  "DIR": 1.0,
  "SPD": 0.07402713881981512,
  "RULE": 1.0,
  "CURV": 0.0,
  "MOM": 0.0,
  "corner_txt": 0.0,
  "corner_prior_term": 0.0,
  "TOTAL": 1.384027138819815,
  "dist": 0.2499999999999959,
  "v": 19.404002087198393,
  "a_eff": 0.0,
  "d_exp": 1.2936001391465595,
  "rel_err": 0.8067408989574391
},
```

Beam Branch value in the intersection

```
{
  "llm_reason": "[Supervisor] The suggested path is straight
                 (theta_deg=0.0) while the driver intends to
                 turn right (TurnIntent=PREP_RIGHT). This
                 contradicts the driver's intent.",
  "DIR": 1.0,
  "SPD": 0.018315650756385214,
  "RULE": 0.0,
  "CURV": 4.2951384375556e-08,
  "MOM": 0.0,
  "corner_txt": 0.3,
  "corner_prior_term": -1.2770640594149771,
  "TOTAL": -0.04874849328141395,
  "dist": 1.0477378640443028e-07,
  "v": 19.404002087198393,
  "a_eff": -8.590276875111201e-09,
  "d_exp": 1.29360013912747,
  "rel_err": 0.999999919006049
},
```

Fig C.2. Fig C.1 presents a comparative visualization of the JSON-formatted outputs for both general roads and intersections. The comparison highlights the structural and informational differences in the system's reasoning when the LLM is utilized versus when it is omitted, illustrating the model's contribution to environmental interpretation.

## APPENDIX D : EVALUATION METHODOLOGY FOR BLIND-SPOT TRAJECTORY

### *D.1. Displacement Errors*

To evaluate physical trajectory accuracy, standard ADE and FDE metrics are employed.

- Average Displacement Error (ADE): The average error across the entire trajectory.

$$ADE = \frac{1}{N}\sum_{i=1}^{N} ||P_i - G_i^*||_2$$

- Final Displacement Error (FDE): The error at the final destination point.

$$FDE = ||P_N - G_N^*||_2$$

## *D.2. Visual Analysis*

In conjunction with quantitative metrics, we visualize the Per-step Distance error trends and Mahalanobis Distance Heatmaps to analyze deviation patterns exhibited by the model in specific scenarios, such as sharp turns or intersection entries.

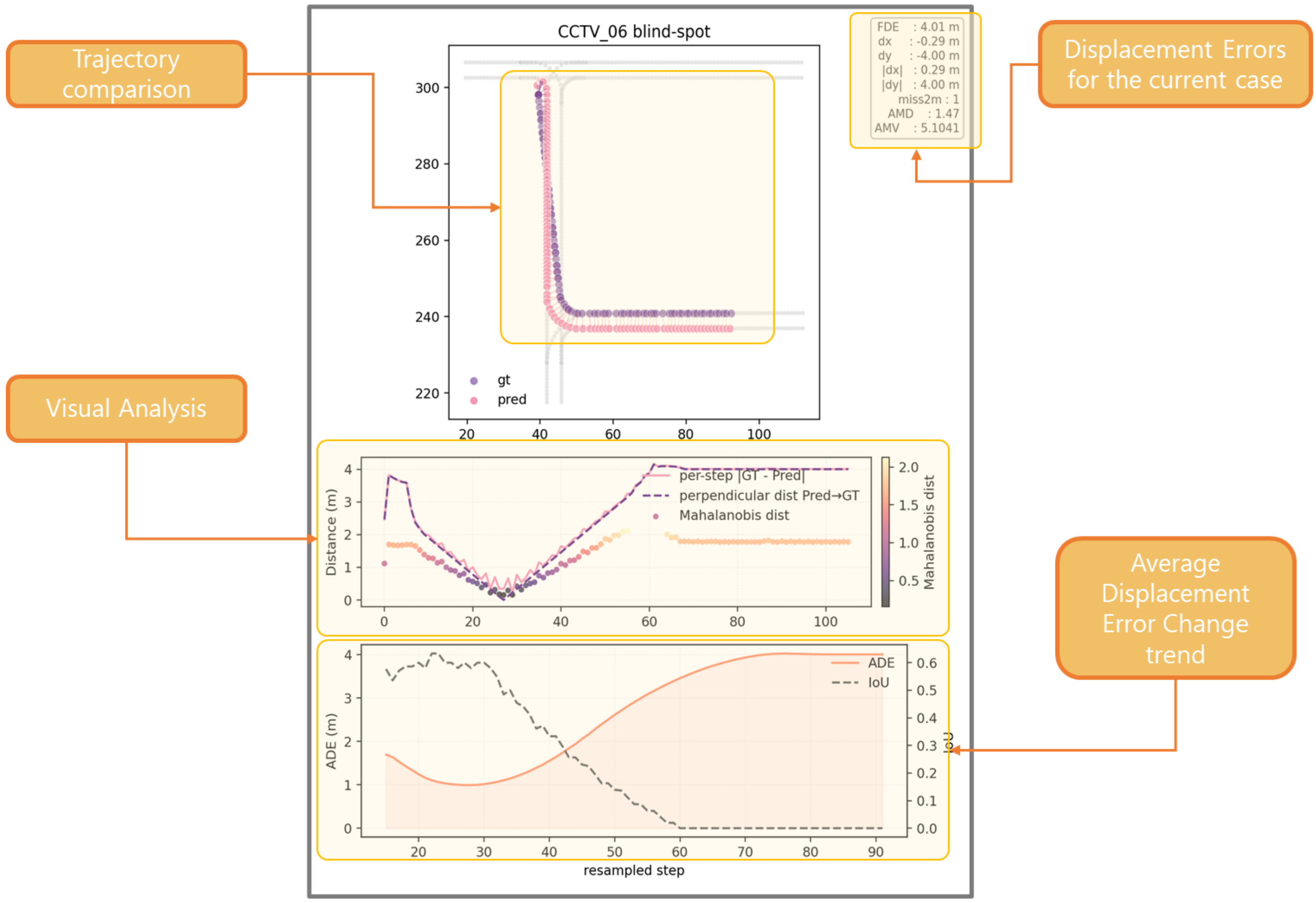

Fig D.1. presents a visual comparison between the actual trajectory of the scenario vehicle (depicted in purple) and the path predicted by Agent 2 (depicted in pink). In the context of this evaluation, the Average Displacement Error (ADE) serves as the sole pertinent metric. This is founded on the premise that the primary objective of this study is the strategic determination of the subsequent CCTV node for handover, rather than achieving high-fidelity alignment between the forecasted path and the ground-truth trajectory.

## APPENDIX E CASE STUDY

### E.1 Agent 1,2 Blind Path Prediction & Next CCTV Selection

GT Trajectory

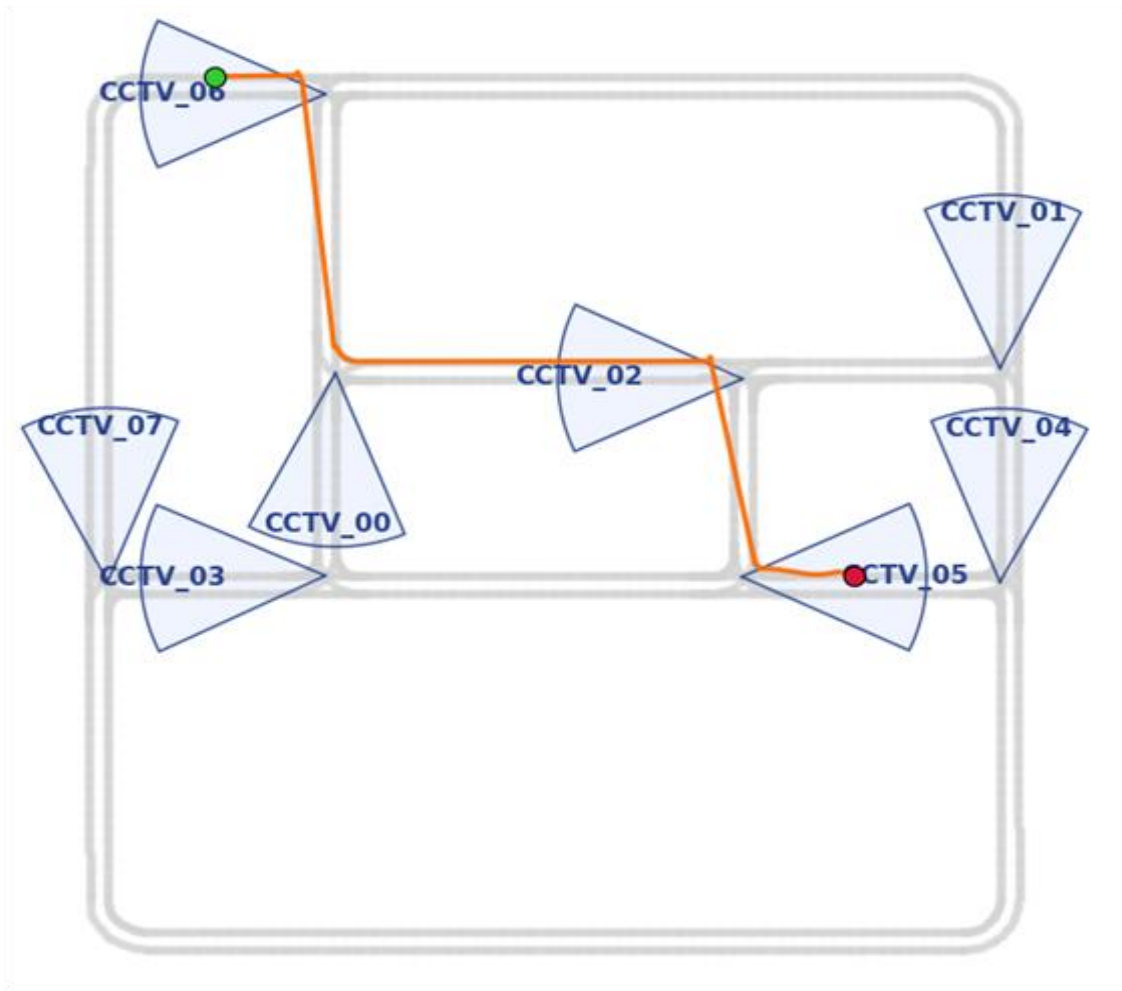

Predicted Trajectory

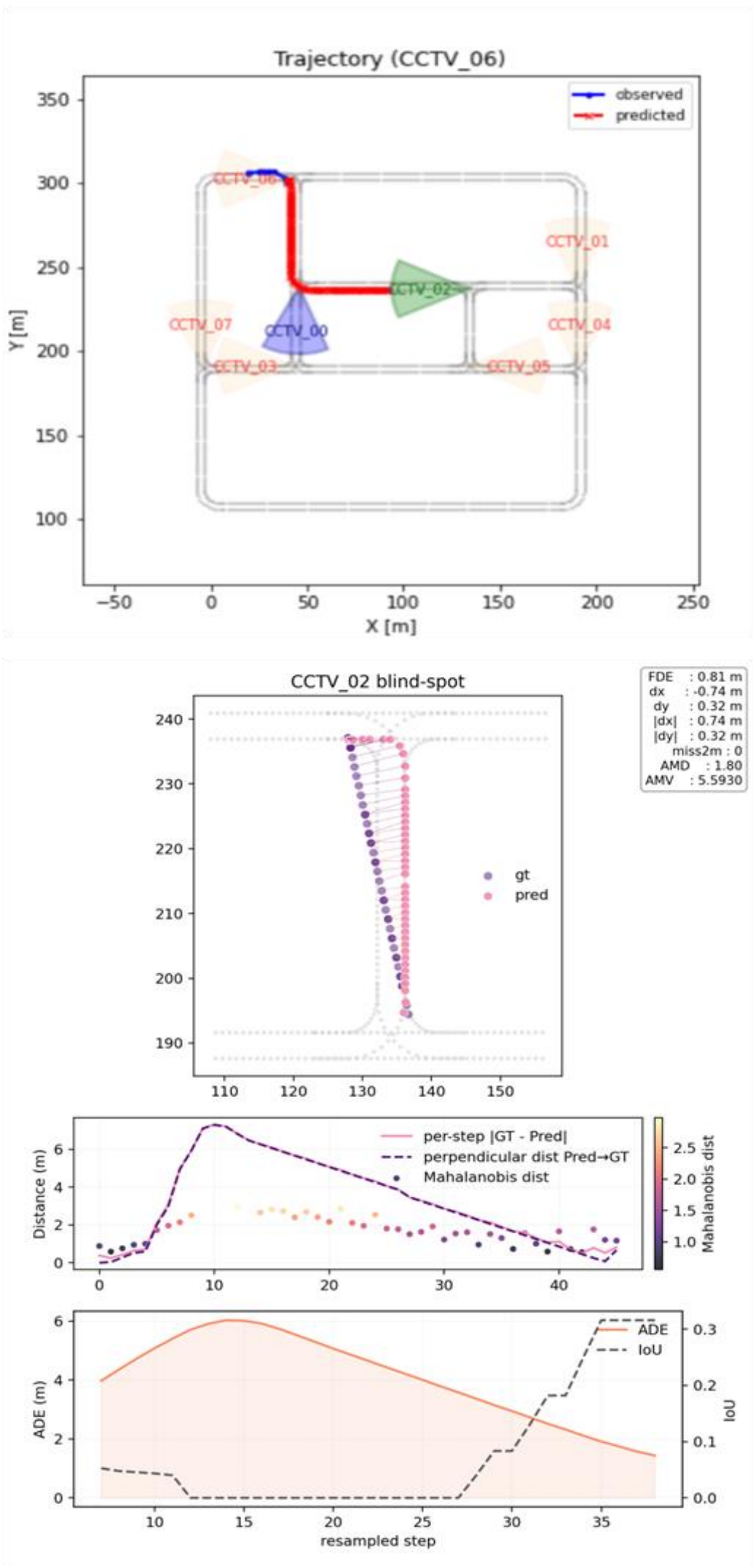

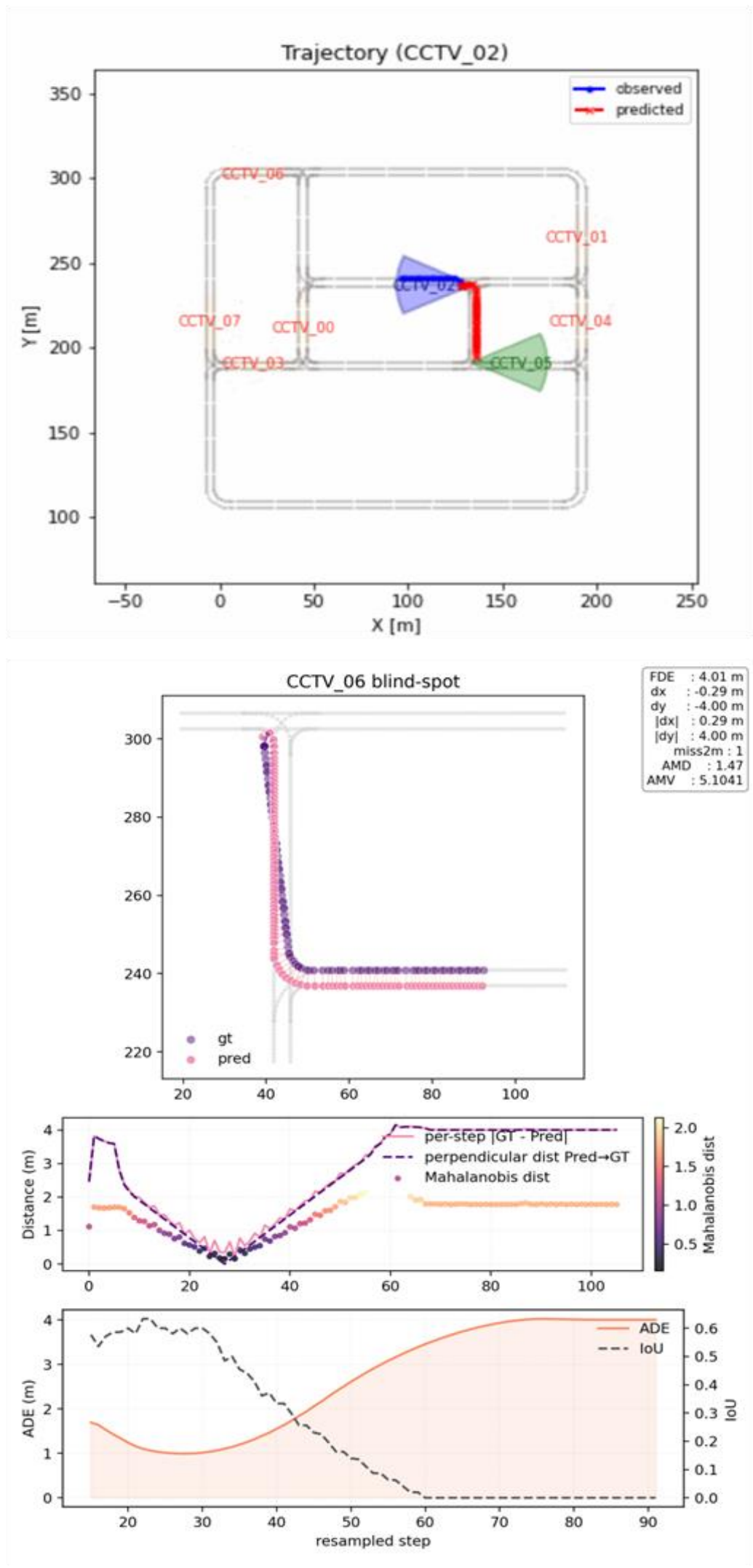

Fig E.1 Scenario 1 illustrates a successful sequential handover. Upon analyzing the trajectory from CCTV_06, the LLM selects CCTV_02 as the primary node (TOP 1, green) among other candidates such as CCTV_00 (blue). The model further demonstrates its predictive accuracy by correctly identifying the subsequent transition from CCTV_02 to CCTV_05, once again assigning CCTV_05 as the TOP 1 candidate.

## GT Trajectory

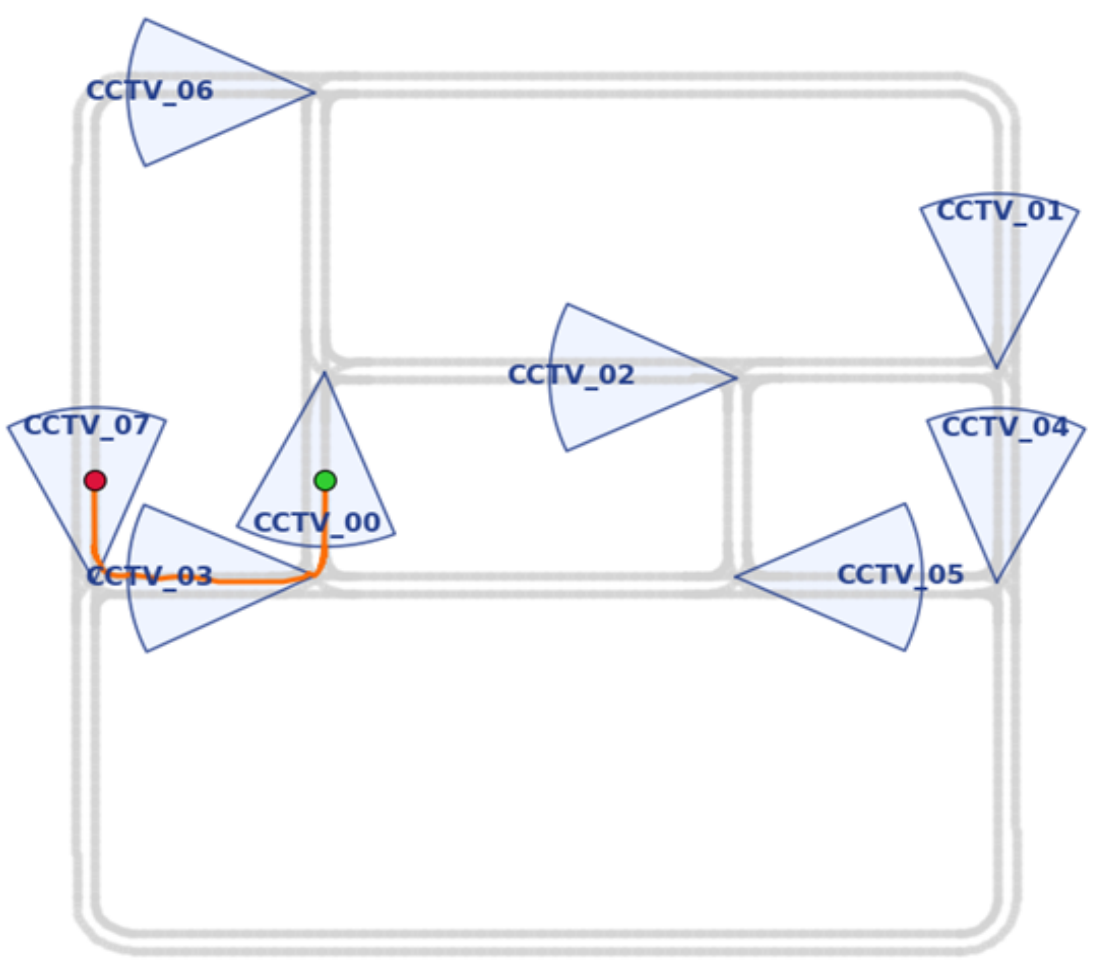

## Predicted Trajectory

Trajectory (CCTV_00)

observed
predicted

X [m]
Y [m]

CCTV_00 blind-spot

gt
pred

FDE : 4.74 m
dx : -0.76 m
dy : -4.68 m
|dx| : 0.76 m
|dy| : 4.68 m
miss2m : 1
AMD : 1.48
AMV : 25.2577

per-step |GT - Pred|
perpendicular dist Pred→GT
Mahalanobis dist
Distance (m)
Mahalanobis dist

ADE
IoU
ADE (m)
IoU
resampled step

Trajectory (CCTV_03)

observed
predicted

X [m]
Y [m]

CCTV_03 blind-spot

gt
pred

FDE : 0.97 m
dx : -0.84 m
dy : -0.48 m
|dx| : 0.84 m
|dy| : 0.48 m
miss2m : 0
AMD : 1.55
AMV : 0.2575

per-step |GT - Pred|
perpendicular dist Pred→GT
Mahalanobis dist
Distance (m)
Mahalanobis dist

ADE
IoU
ADE (m)
IoU
resampled step

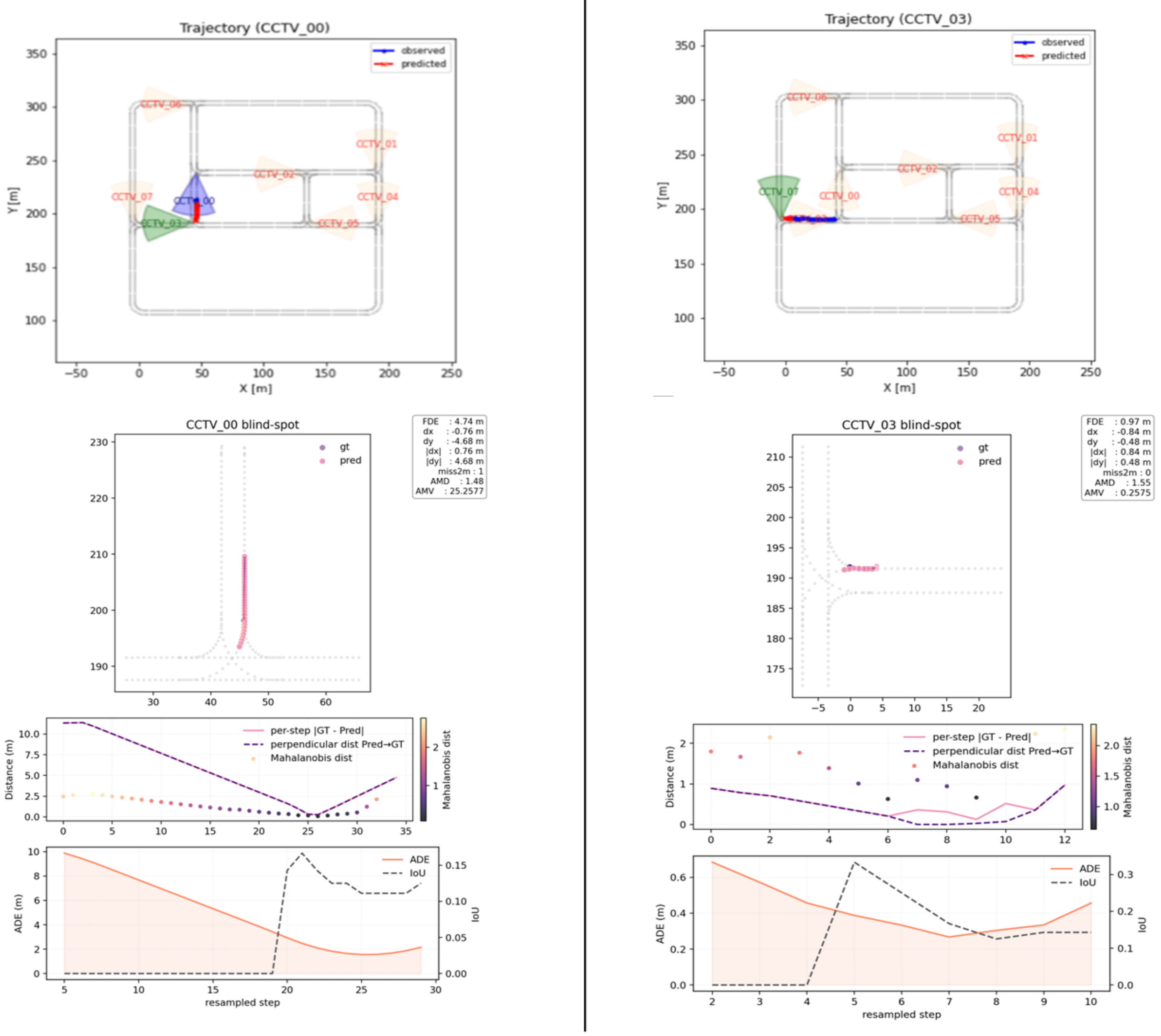

Fig E.2. This scenario illustrates a sequential transition starting from CCTV_00, passing through CCTV_03, and concluding at CCTV_07. Notably, CCTV_00 was included in the candidate set despite the observation concluding mid-trajectory within its range. This inclusion is attributed to the system's selection criterion, which mandates that a target must be observed within a camera's Field of View (FOV) for at least 15 time-steps to be registered as a valid candidate. Since the vehicle's presence within CCTV_00's FOV exceeded this temporal threshold before the tracking sequence ended, the node was correctly identified and maintained within the candidate pool.

## GT Trajectory

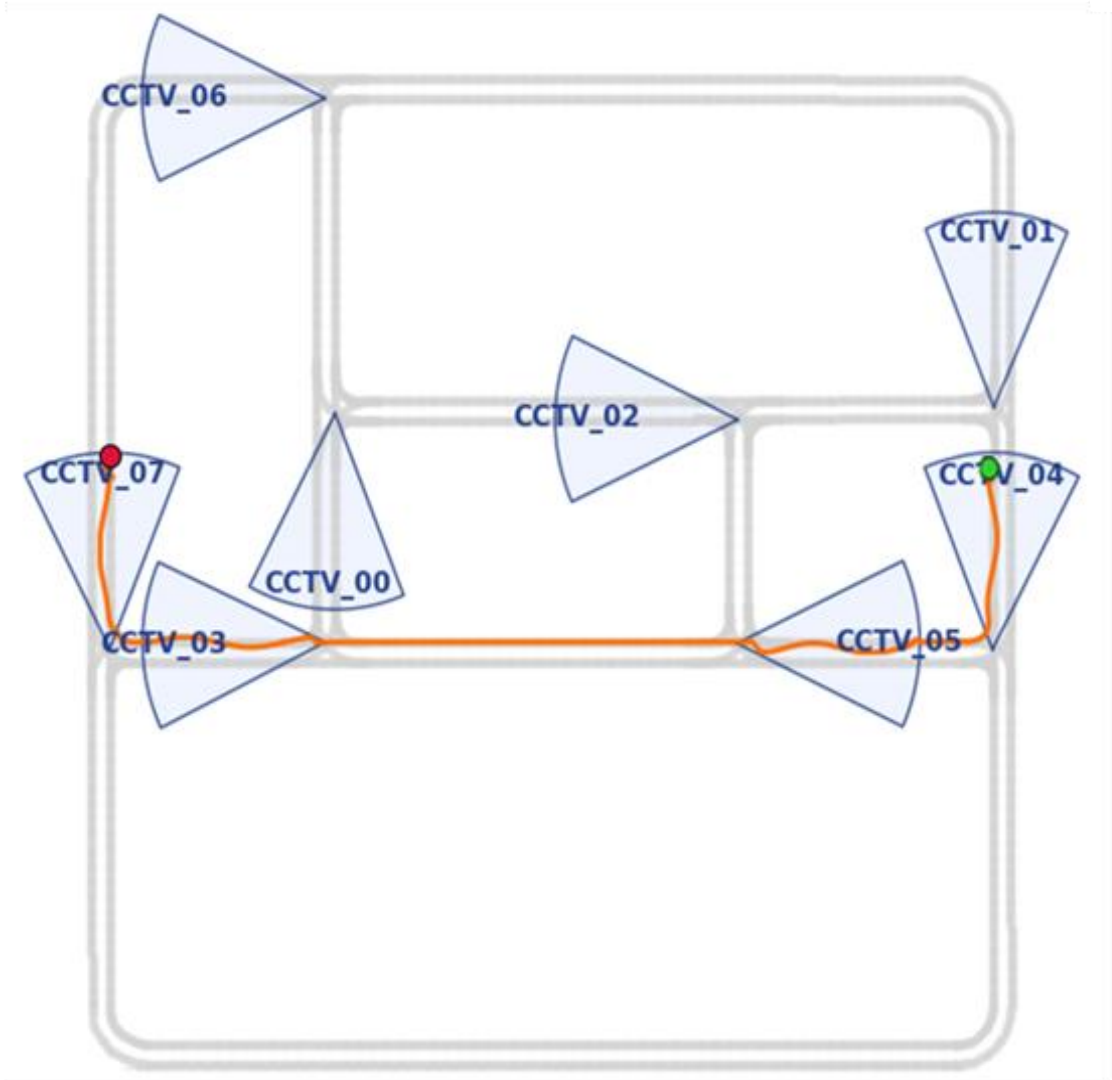

## Predicted Trajectory

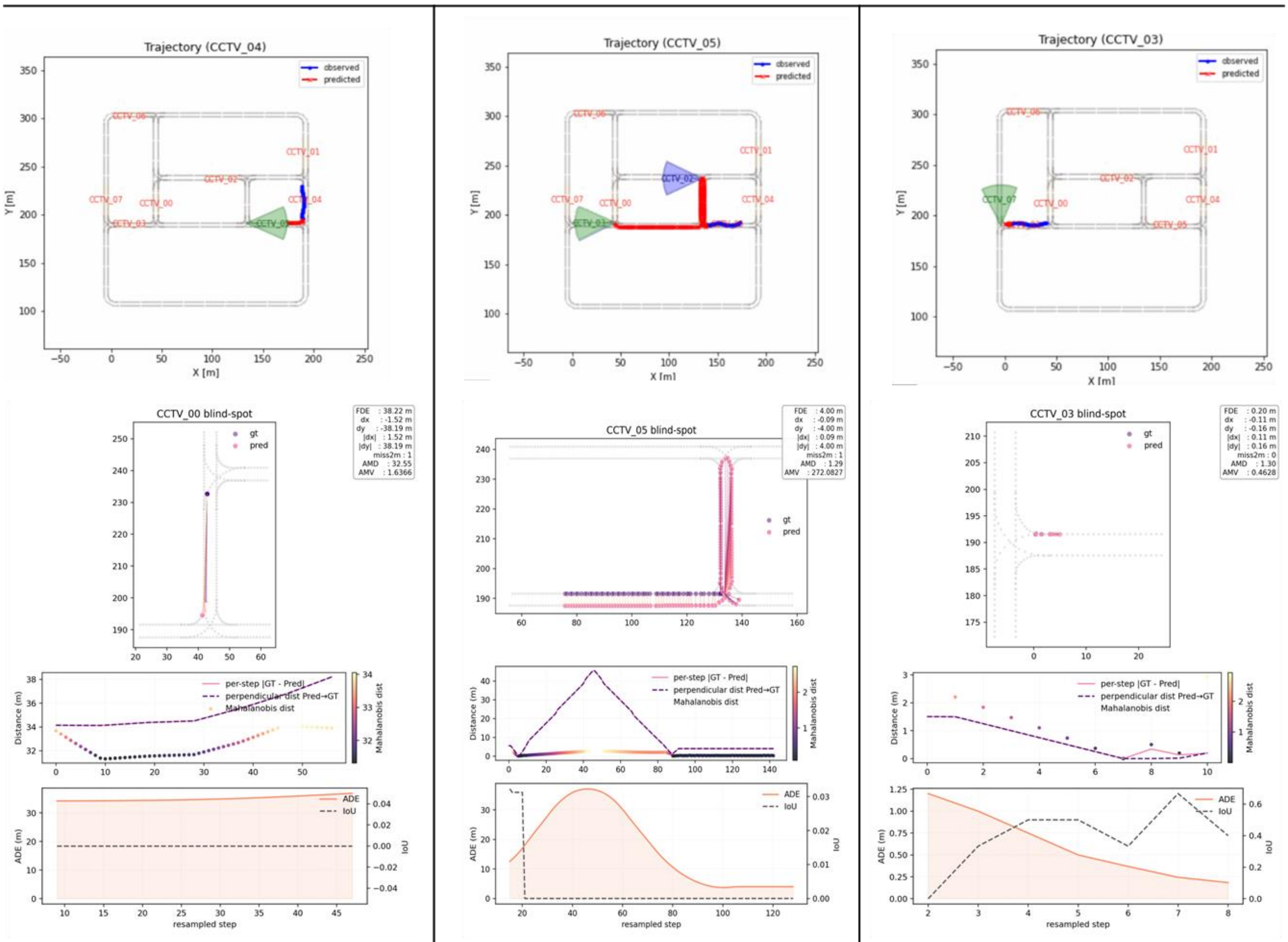

Fig E.3 This case presents a high-complexity scenario involving a sinuous trajectory within the coverage of CCTV_04. While the system successfully anticipated the transition to CCTV_05, a significant challenge arose as the vehicle exited the Field of View (FOV): the target exhibited a rightward heading followed by a brief linear displacement. This behavioral shift introduced spatial ambiguity, leading the model to initially include CCTV_02 in the candidate set. However, by assigning a higher weight to the terminal directional intent, the orchestrator correctly prioritized CCTV_03 as the most probable destination. The subsequent successful acquisition at CCTV_07 validates the effectiveness of the proposed scoring heuristic in resolving directional uncertainty during FOV transitions.

*E.2 Response comparison among LLMs per case*

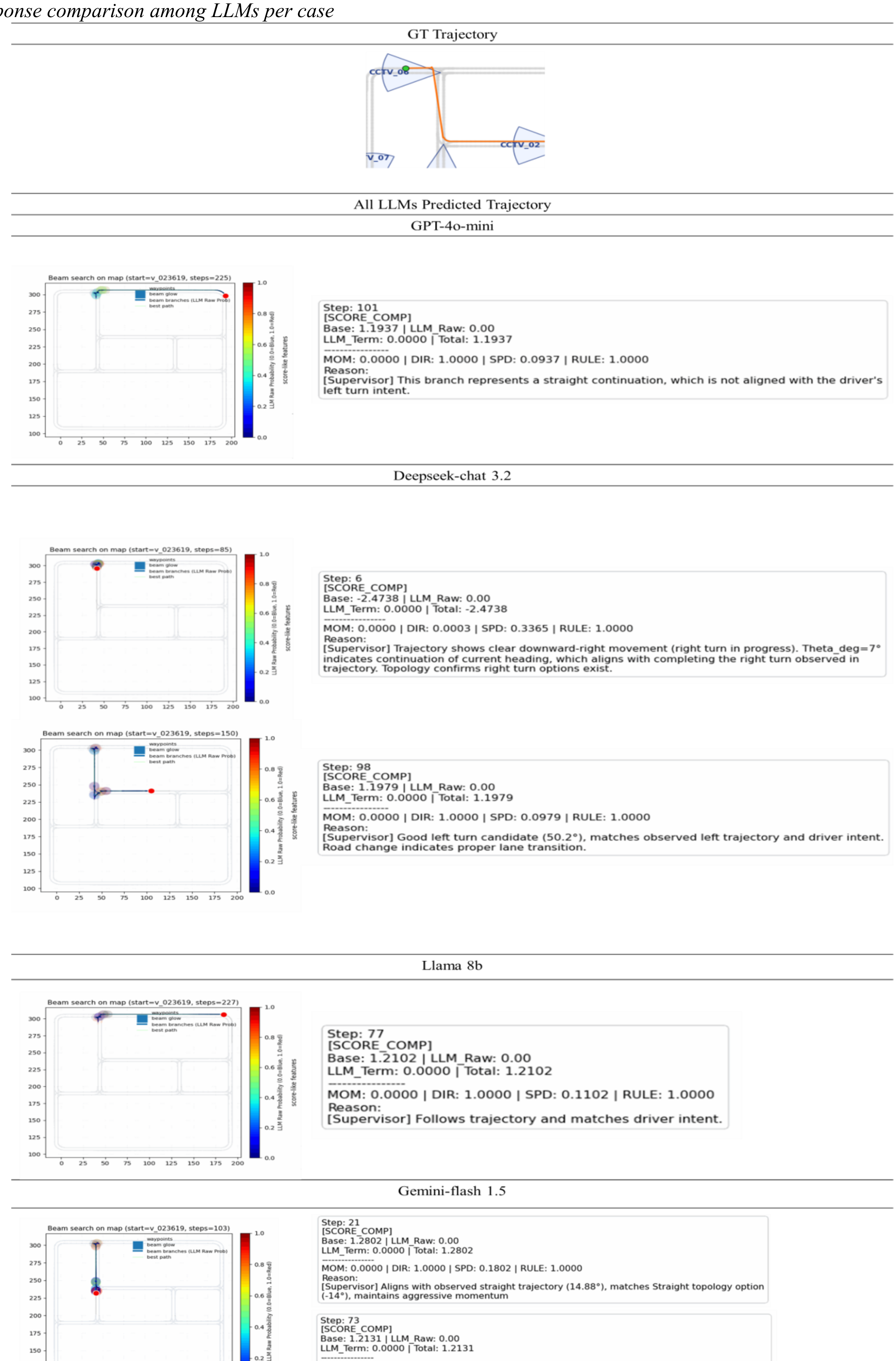

Fig E.4. This section provides a comparative analysis of how different LLMs respond to the same high-complexity scenario. Among the evaluated models, DeepSeek was the sole architecture to successfully navigate the spatial ambiguities and yield the correct prediction. This outcome suggests that DeepSeek possesses a superior capacity for nuanced intent analysis, whereas other models failed to reconcile the conflicting directional cues at the point of FOV exit.

GT Trajectory

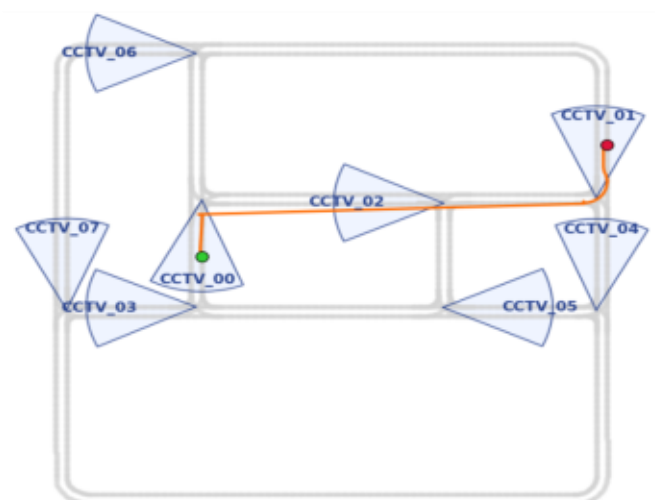

All LLMs Predicted Trajectory

GPT-4o-mini

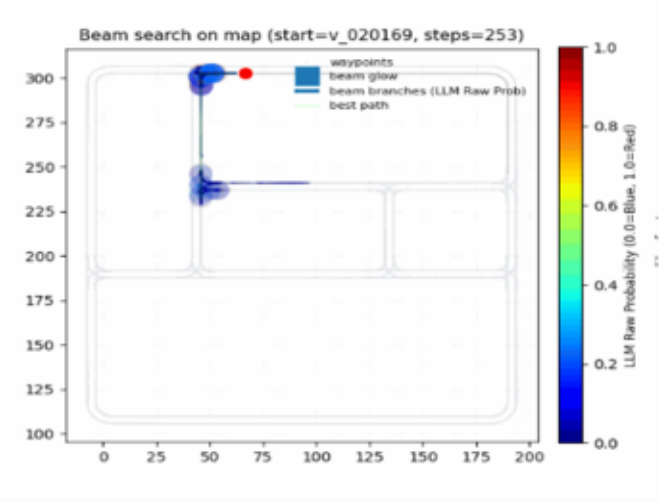

Step: 13
[SCORE_COMP]
Base: 2.0194 | LLM_Raw: 0.00
LLM_Term: 0.0000 | Total: 2.0194
----------------
MOM: 0.0000 | DIR: 0.9969 | SPD: 0.9993 | RULE: 1.0000
Reason:
[Supervisor] The trajectory is straight ahead and matches the left turn intent, indicating a strong continuity.

Step: 76
[SCORE_COMP]
Base: 1.3131 | LLM_Raw: 0.00
LLM_Term: 0.0000 | Total: 1.3131
----------------
MOM: 0.0000 | DIR: 1.0000 | SPD: 0.2131 | RULE: 1.0000
Reason:
[Supervisor] This branch represents a left turn, which aligns with the driver's high left turn intent and is appropriate given the intersection context.

Deepseek-chat 3.2

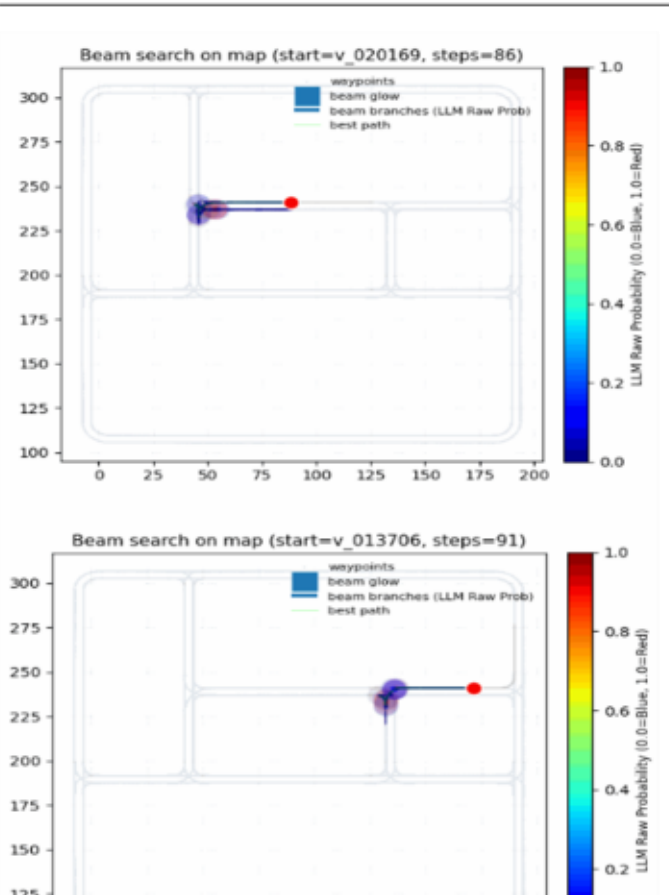

Step: 42
[SCORE_COMP]
Base: 1.4622 | LLM_Raw: 0.00
LLM_Term: 0.0000 | Total: 1.4622
----------------
MOM: 0.0000 | DIR: 1.0000 | SPD: 0.3622 | RULE: 1.0000
Reason:
[Supervisor] Best available turn option (89.94°). Vehicle is approaching (not turning), driver intent strongly Left (0.96). This is the only significant turn branch that could represent the left turn in this T-intersection context.

Step: 36
[SCORE_COMP]
Base: 2.2421 | LLM_Raw: 0.00
LLM_Term: 0.0000 | Total: 2.2421
----------------
MOM: 0.5000 | DIR: 1.0000 | SPD: 0.3171 | RULE: 1.0000
Reason:
[Supervisor] Strong continuity. Theta (16.19°) continues the rightward turn, matching the active maneuver and the intersection's 'Sharp Right' option. Momentum_match is 1.

Llama 8b

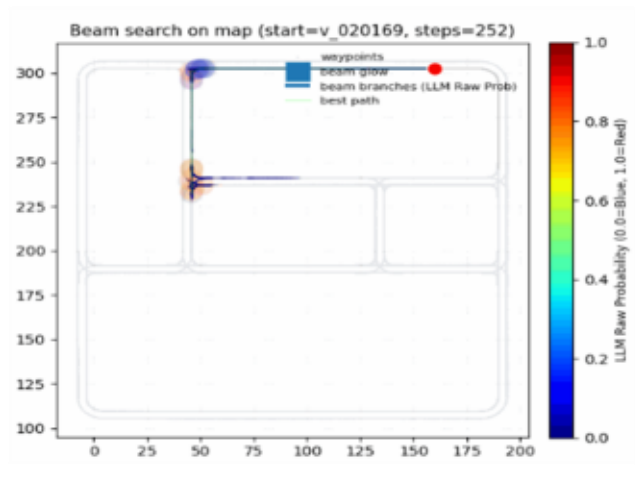

Step: 7
[SCORE_COMP]
Base: 1.6219 | LLM_Raw: 0.00
LLM_Term: 0.0000 | Total: 1.6219
----------------
MOM: 0.0000 | DIR: 0.9889 | SPD: 0.6731 | RULE: 1.0000
Reason:
[Supervisor] Strong match with Intent to turn Left and active turn to the Left.

Step: 75
[SCORE_COMP]
Base: 1.3145 | LLM_Raw: 0.00
LLM_Term: 0.0000 | Total: 1.3145
----------------
MOM: 0.0000 | DIR: 1.0000 | SPD: 0.2145 | RULE: 1.0000
Reason:
Perfect match with high-score Intent (Left=0.96)

Gemini-flash 1.5

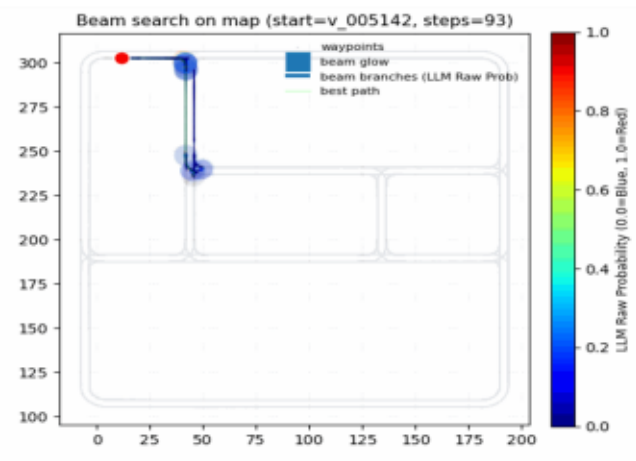

Step: 15
[SCORE_COMP]
Base: 1.6779 | LLM_Raw: 0.00
LLM_Term: 0.0000 | Total: 1.6779
----------------
MOM: 0.0000 | DIR: 1.0000 | SPD: 0.5779 | RULE: 1.0000
Reason:
[Supervisor] This branch aligns with the active left turn maneuver and high turn intent.

Step: 86
[SCORE_COMP]
Base: 1.8817 | LLM_Raw: 0.00
LLM_Term: 0.0000 | Total: 1.8817
----------------
MOM: 0.0000 | DIR: 1.0000 | SPD: 0.7817 | RULE: 1.0000
Reason:
[Supervisor] Strong match for left turn; trajectory is aligned with the left turn intent.

Fig E.5 Second example

APPENDIX F EXPERIMENTAL ENVIRONMENT SET-UP DETAIL

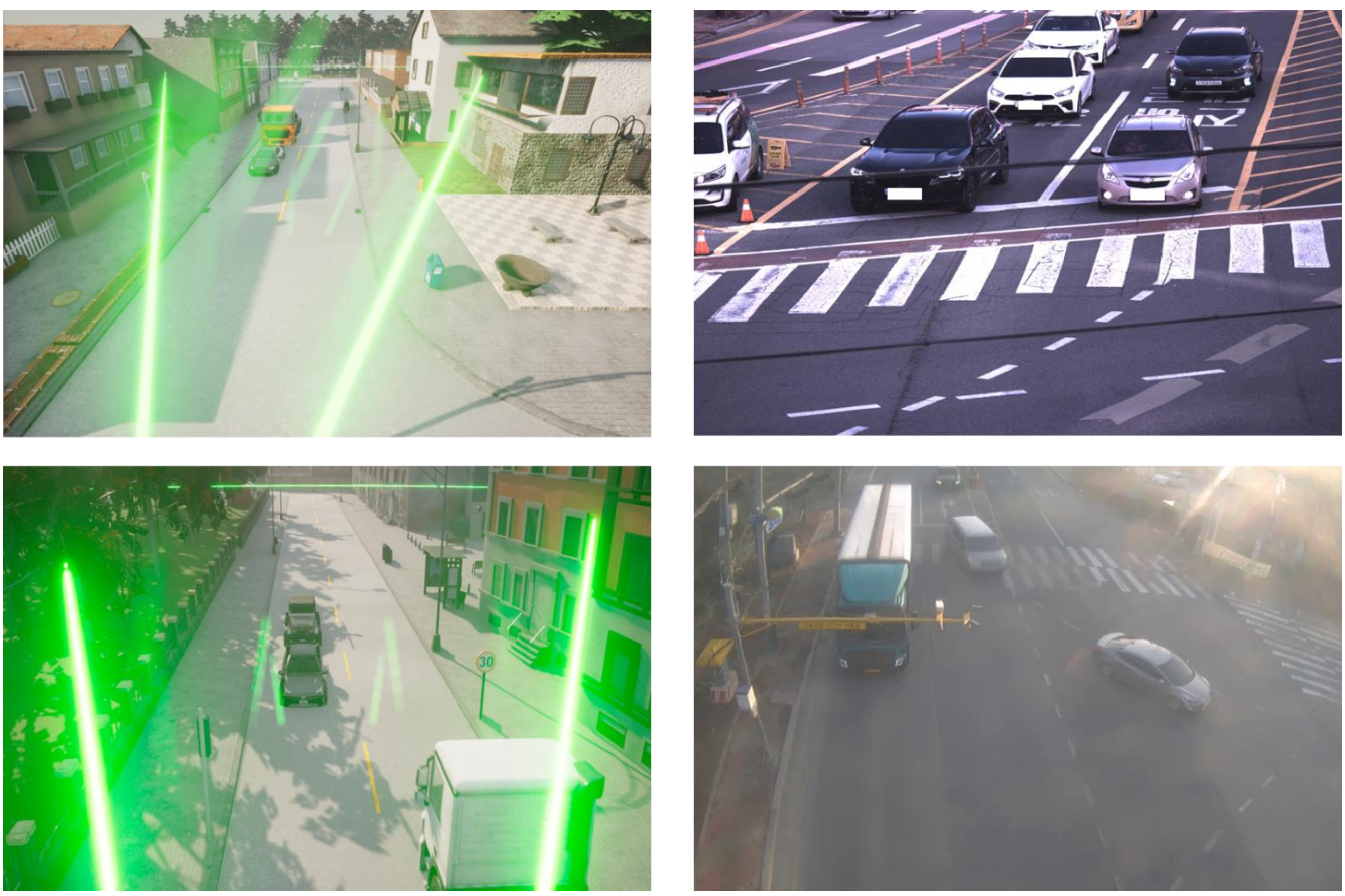

Fig F.1. The simulation environment exhibits a high degree of fidelity to actual conditions. While vehicle dimensions in the CARLA environment (left) are marginally smaller than those in the real-world footage, and light saturation is observed in the bottom-right quadrant of the real-world data, these discrepancies have a negligible impact on empirical validity as such saturated regions are typically excluded from analysis in practical applications.

### *F.1. Infrastructural Consistency and Regulatory Compliance*

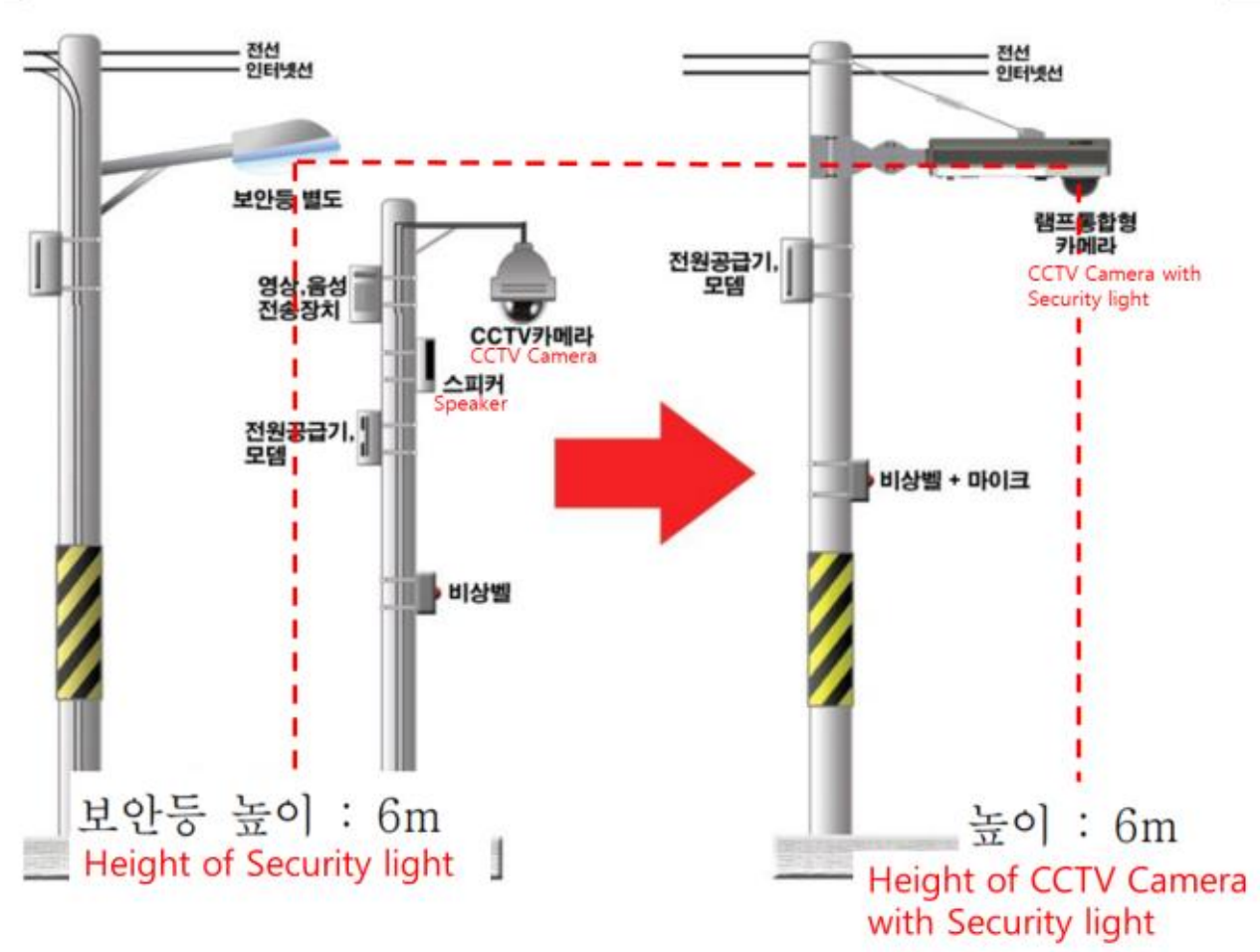

Fig F.1 The installation and operational parameters of the CCTV system were established in accordance with SSF-ST-O-0005 (Standard for CCTV System Operation and Maintenance), an association standard enacted by the Smart City Standardization Forum on January 18, 2013. This ensures that the simulation environment adheres to the recognized technical protocols for urban surveillance infrastructure.

The deployment of virtual CCTV units within the CARLA environment—positioned at a height of 6.0 m with a -20.0° pitch—rigidly adheres to the "Manual on Traffic Signal Installation and Management" published by the Korean National Police Agency.

- Vertical Clearance: In accordance with the manual, the bottom of overhanging (side-post horizontal) traffic signals must maintain a minimum vertical clearance of 4.5 m (450 cm) from the roadway surface.

- Optimal Perspective: By elevating the sensors to 6.0 m, this study effectively replicates real-world urban infrastructure while minimizing visual occlusions caused by oversized vehicles, such as heavy-duty trucks. This configuration ensures that the LLM agent operates on realistic vertical Field of View (FOV) data, providing a high-fidelity surrogate for actual urban monitoring conditions.

*F.2.* Advantages of Simulation-based Evaluation

While real-world testing is essential, it presents significant limitations regarding safety, operational costs, and the difficulty of constructing reproducible "Hard Scenarios." The simulation environment offers several critical advantages for validating the proposed framework:

- Ground-Truth Precision: The system facilitates the acquisition of high-fidelity world coordinates $(x, y, z)$ via the gt_traj.csv dataset. This allows for a precise quantitative assessment of the Average Displacement Error (ADE), serving as a rigorous benchmark for the predictive algorithms.

- Scenario Diversity: The simulation allows for the safe and iterative reproduction of complex maneuvers that necessitate Beam Search and LLM-based intent analysis. This includes sinuous trajectories and high-uncertainty blind-spot transitions that are difficult to capture in spontaneous real-world traffic.

- Temporal Thresholding: The controlled environment enables the optimization of data preprocessing heuristics. Specifically, a temporal threshold of 15 observation steps was established within the FOV to filter out transient noise and ensure the LLM receives sufficiently contextualized trajectory segments for reasoning.